\documentclass[myepj-spec,pdftex, iicol]{mySvjour}
\usepackage{graphicx}
\usepackage[pdfpagemode=UseNone]{hyperref}
\usepackage{color}
\usepackage{xspace}
\usepackage[square,sort&compress,numbers]{natbib}   
\usepackage{verbatim}
\usepackage{amsmath,amssymb,amsfonts} 
\usepackage{tabularx}
\usepackage{multicol}
\usepackage{footnote}
\usepackage{listings}
\usepackage[gen]{eurosym}
\usepackage[switch, modulo]{lineno}
\usepackage{longtable}
\usepackage[utf8]{inputenc} 
\usepackage[T1]{fontenc}    
\usepackage{hyperref}       
\usepackage{url}            
\usepackage{booktabs}       
\usepackage{amsfonts}       
\usepackage{nicefrac}       
\usepackage{microtype}      
\usepackage{xcolor}         
\usepackage{graphicx}
\usepackage{float}
\usepackage{caption}
\usepackage{subcaption}
\usepackage{amsmath}

\usepackage{lmodern,bm}                
\usepackage[T1]{sansmath} 
\SetMathAlphabet{\mathsfbf}{sans}{\sansmathencoding}{\sfdefault}{bx}{sl}
\usepackage{etoolbox}
\AtBeginEnvironment{sansmath}{}{}{}

\usepackage{lipsum}

\definecolor{darkblue1}{rgb}{0,0,.2}
\definecolor{darkblue}{rgb}{0,0,.2}
\definecolor{darkred}{rgb}{0.5,0,0}
\pagecolor{white} 
\color{black}     
\hypersetup{breaklinks=true, 
	colorlinks=true, 
	linkcolor=darkblue1, 
	menucolor=darkblue1, 
	urlcolor=darkblue1,
	citecolor=darkblue1,
	pdftitle={},
	pdfauthor={},
	pdfsubject={},
	pdfkeywords={},
	pdfproducer={}
}
\usepackage{siunitx}

\bibstyle{plain}
\begin{document}

			\begin{flushright}
				\normalsize
			\end{flushright}
			
			\vspace{-2cm}
			
			\title{\Large\boldmath Uncovering Hidden Systematics in Neural Network Models for High Energy Physics}

\author{Lucie Flek\inst{1} \and Philipp Alexander Jungs\inst{2} \and Akbar Karimi\inst{1} \and Timo Saala\inst{3} \and Alexander Schmidt\inst{2} \and Matthias L. Schott\inst{3}\thanks{\emph{Corresponding author(s). E-mail(s):} \href{mailto:mschott@uni-bonn.de}{mschott@uni-bonn.de}} \and Philipp Soldin\inst{4} \and Christopher Wiebusch\inst{4} \and Ulrich Willemsen\inst{2}%
}                     
\offprints{}          
\institute{Bonn-Aachen Institute of Technology, University of Bonn, Germany \and Physics Institute III A, RWTH Aachen University, Germany \and Institute of Physics, University of Bonn, Germany \and Physics Institute III B, RWTH Aachen University, Germany }

			
\abstract{Neural networks (NNs) are inherently multidimensional classifiers that learn complex, non-linear relationships among input observables. While their flexibility enables unprecedented performance in high-energy physics (HEP) analyses, it also makes them sensitive to small variations in their inputs. Consequently, the propagation and estimation of systematic uncertainties in NN-based models remain an open challenge.
There are indications that uncertainties derived in control regions or from nominal variations of input features can underestimate the true model uncertainty, potentially leaving biases unaccounted for. Inspired by insights from adversarial-attack studies in machine learning, we explore how subtle perturbations, fully consistent with the experimental uncertainties on the input observables, can lead to substantial changes in NN outputs, while keeping the one-dimensional and correlated input distributions nearly unchanged. Using a set of representative HEP tasks, including event classification and object identification, and testing across a variety of network architectures, we demonstrate that networks can be systematically “fooled” at significant rates within the allowed uncertainty envelopes. Building on this observation, we introduce a quantitative framework to probe and measure the hidden sensitivity of neural networks to realistic experimental variations, providing a practical path to evaluate and control their systematic uncertainty in physics analyses.}

\maketitle

\tableofcontents

\section{Introduction and Related Work}
\label{sec:introduction}

Deep neural networks (DNNs) have become a central component of modern analyses at the Large Hadron Collider (LHC), enabling powerful discrimination between signal and background processes as well as improved reconstruction of complex observables~\cite{Feickert:2021ajf,Plehn:2022ftl}. Their performance gains over traditional cut-based or shallow multivariate methods arise from their ability to exploit high-dimensional correlations among input features. In typical high-energy physics (HEP) applications, DNNs are trained on Monte Carlo (MC) simulated samples, under the assumption that the simulations faithfully reproduce the relevant correlations present in nature. Validation procedures rely primarily on comparisons of one-dimensional (1D) input feature distributions between data and simulation in control regions, occasionally supplemented by selected higher-dimensional checks. Residual mismodeling effects are incorporated through systematic uncertainties derived from variations of MC generators, parton shower models, hadronization schemes, and detector response parameters~\cite{Nachman:2020guide,Ghosh:2021review}.

This standard workflow implicitly assumes that agreement in validated observables guarantees reliable classifier behavior in the full feature space. While this assumption has been sufficient for traditional low-capacity models, its validity for modern high-dimensional architectures is less clear. In particular, DNNs may learn decision boundaries that depend on complex, nonlinear correlations that are not explicitly constrained by standard validation strategies.

In the broader machine-learning literature, it is well established that high-capacity classifiers are vulnerable to adversarial perturbations. Methods such as DeepFool~\cite{MoosaviDezfooli:2016deepfool} and projected gradient descent (PGD) attacks~\cite{Madry:2017pgd} demonstrate that small, structured perturbations, often imperceptible in low-dimensional projections, can significantly alter model predictions. While originally studied in computer vision, the geometric mechanism underlying adversarial vulnerability, the sensitivity of complex decision boundaries to coherent deformations in high-dimensional space, applies generically to deep classifiers.

Robustness and uncertainty quantification in HEP have been addressed from complementary perspectives. Inference-aware approaches such as INFERNO~\cite{deCastro:2018inferno} directly incorporate systematic variations into the training objective. Uncertainty-aware learning strategies emphasize consistent treatment of nuisance parameters during optimization~\cite{Ghosh:2021robustness,Bellagente:2021cautionary}. Bayesian and ensemble-based methods have been proposed to quantify predictive uncertainties in jet tagging and related tasks~\cite{Bollweg:2020bnnjets,Dillon:2024ensembles,Aparicio:2024evidential}. Practical deployment guidelines further stress the need to account for systematic effects when using deep learning in LHC searches~\cite{Nachman:2020guide}, and recent experimental applications have begun integrating uncertainty-aware neural networks into calibration and reconstruction workflows~\cite{ATLAS:2024bnncalib,Gavrikov:2024acorn}. 

A complementary line of investigation explores adversarial robustness directly in the context of physics analyses. Several studies have shown that substantial fooling ratios can be achieved in event classification tasks while leaving all 1D feature distributions unchanged, and even while preserving linear correlations or decorrelation constraints \cite{Flek:2024minifool, Flek:2025ecg, Bechtle:2026nlc}. These findings highlight that classifier behavior can change significantly without being detectable through standard validation plots. However, unconstrained adversarial attacks may exploit directions in feature space that are mathematically allowed but physically implausible, corresponding to deformations that would never arise from realistic detector or modeling effects. As a result, the interpretation of such fooling ratios as genuine sources of uncertainty in physics measurements remains ambiguous.

An important step toward physical realism was introduced with the MiniFool framework~\cite{Flek:2024minifool}, which constrains adversarial perturbations to lie within experimentally motivated uncertainty bounds on input features. By construction, this ensures compatibility with detector resolution effects. Nevertheless, MiniFool permits coherent, one-sided shifts of features across the entire sample, leading to visible distortions of 1D distributions and raising concerns regarding physical plausibility. Realistic experimental and modeling uncertainties are expected to manifest as approximately Gaussian-distributed fluctuations centered around zero, without a preferred direction in feature space. The existing literature therefore leaves a conceptual gap between purely adversarial constructions and physically meaningful uncertainty estimates for DNN-based classifiers in HEP. While it has been demonstrated that classifiers can be fooled under increasingly stringent constraints, it remains unclear whether such effects correspond to realistic variations that should be treated as systematic uncertainties in physics analyses.

In this work, we bridge this gap by constructing adversarial perturbations that are simultaneously constrained by experimental uncertainties, implicitly preserve low-dimensional and higher-order correlations, and remain statistically indistinguishable from nominal samples under standard validation procedures. Specifically, we minimize the $\chi^2$ distance between nominal and perturbed 1D feature distributions while enforcing zero-mean Gaussian-distributed feature deviations. The resulting adversarial samples satisfy conventional validation criteria, including stability of 1D projections and linear correlation matrices, yet still induce measurable performance shifts in high-capacity classifiers. As a complementary perspective, we also investigate retraining strategies in which classifiers are trained on mixtures of nominal and adversarially perturbed samples. The performance difference between nominally trained and retrained models, evaluated on independent nominal test data, provides an estimate of a model-dependent uncertainty intrinsic to high-capacity learning algorithms.

\section{Conceptual Framework and Uncertainty-Constrained Adversarial Construction}
\label{sec:framework_attack}

In this work, we quantify the sensitivity of high-capacity classifiers to physically plausible perturbations of their input features.
We define the \emph{fooling ratio} as the fraction of events for which a classifier prediction changes under a constrained perturbation of the input.
This metric directly probes the stability of the learned decision boundary and is independent of the overall classification accuracy.
A non-zero fooling ratio indicates that small, experimentally allowed deformations of the input space can induce qualitative changes in classifier decisions.

To ensure physical relevance, perturbations are required to satisfy the following criteria:
(i) they must remain within experimentally motivated uncertainties of the input features,
(ii) they must not induce observable distortions in one-dimensional feature distributions,
(iii) they must exhibit no preferred direction in feature space, consistent with Gaussian-distributed fluctuations,
and (iv) they must preserve correlations to all practically accessible orders.
These requirements guarantee that the perturbed samples remain statistically indistinguishable from nominal events under standard validation procedures.

Adversarial examples are constructed using a white-box projected gradient descent (PGD) procedure subject to these constraints.
Given a trained classifier $f_\theta(\mathbf{x})$ with parameters $\theta$, input features $\mathbf{x} \in \mathbb{R}^D$, and labels $y$, we seek a perturbed input $\mathbf{x}^{\mathrm{adv}}$ that modifies the classifier prediction while remaining within the physically allowed region.

For each feature $x_i$, we define a Gaussian uncertainty model
\begin{equation}
\sigma_i = f_i \, |x_i|,
\end{equation}
where $f_i$ denotes the fractional uncertainty associated with feature $i$.
This reflects the approximate scaling of many detector uncertainties with the magnitude of the observable.
The perturbation is defined as
\begin{equation}
\Delta_i = x_i^{\mathrm{adv}} - x_i,
\end{equation}
and is restricted to lie within a fixed number of standard deviations,
\begin{equation}
|\Delta_i| \leq N_\sigma \, \sigma_i,
\end{equation}
where $N_\sigma$ controls the maximal allowed deviation.

The perturbation is obtained by maximizing the composite objective function
\begin{equation}
\label{eqn:crossattack}
\mathcal{L}_{\mathrm{attack}}
=
\mathcal{L}_{\mathrm{CE}}(f_\theta(\mathbf{x}^{\mathrm{adv}}), y)
-
\lambda_{\chi^2} \, \chi^2(\mathbf{x}, \mathbf{x}^{\mathrm{adv}})
-
\lambda_{\Delta} \, \mathcal{L}_{\mathrm{prior}}(\mathbf{x}, \mathbf{x}^{\mathrm{adv}}).
\end{equation}
The first term is the cross-entropy loss, which is maximized in order to induce misclassification.
The second term penalizes deviations in one-dimensional feature distributions via a soft $\chi^2$ measure computed over binned observables, thereby suppressing distortions that would be visible in control plots.
The third term enforces statistical realism through a Gaussian prior on the perturbations.

Defining standardized deviations
\begin{equation}
z_i = \frac{\Delta_i}{\sigma_i},
\end{equation}
the Gaussian prior is implemented as
\begin{equation}
\mathcal{L}_{\mathrm{prior}}
=
\left\langle z_i^2 \right\rangle_{i,\mathrm{batch}},
\end{equation}
which penalizes coherent, one-sided shifts and favors perturbations distributed symmetrically around zero, as expected for experimental fluctuations.

The adversarial input is constructed iteratively using projected gradient ascent,
\begin{equation}
\mathbf{x}^{(t+1)} =
\Pi_{\mathcal{C}}
\left[
\mathbf{x}^{(t)}
+
\alpha \,
\mathbf{w}(\mathbf{z}) \odot 
\mathrm{sign}
\left(
\nabla_{\mathbf{x}} \mathcal{L}_{\mathrm{attack}}
\right)
\right],
\end{equation}
where $\alpha$ denotes the step size, $\odot$ represents element-wise multiplication, and $\Pi_{\mathcal{C}}$ projects onto the feasible region defined by the per-feature bounds and optional global limits.
The update is modulated by a Gaussian weight
\begin{equation}
w_i(z_i) = \exp\left(-\frac{1}{2} z_i^2\right),
\end{equation}
which suppresses large relative deviations and further concentrates the perturbations in the high-probability region of the Gaussian prior \footnote{The fact that the allowed perturbations are small implicitly also ensures that correlations are preserved to a large degree.}.
After each iteration, the input is projected onto the allowed interval
\begin{equation}
x_i^{\mathrm{adv}} \in 
\left[x_i - N_\sigma \sigma_i,\; x_i + N_\sigma \sigma_i \right],
\end{equation}
ensuring strict compliance with the experimental uncertainty model.

By construction, the resulting adversarial samples (i) induce changes in classifier decisions, (ii) preserve one-dimensional feature distributions, (iii) respect experimentally motivated uncertainty bounds, and (iv) exhibit Gaussian-distributed, zero-mean perturbations without a preferred direction.
They therefore represent physically plausible deformations of the input space that evade standard validation techniques while probing latent sensitivities of high-capacity classifiers.

As an alternative to the cross-entropy-based objective, we also consider a loss function inspired by the optimization framework introduced by Carlini and Wagner (C\&W)~\cite{DBLP:journals/corr/CarliniW16a}, extended to incorporate the physical constraints introduced in the previous section. In this hybrid formulation, adversarial perturbations are obtained by minimizing a composite objective that simultaneously penalizes the size of the perturbation, enforces misclassification through a margin-based criterion, and constrains the perturbed sample to remain statistically consistent with the nominal distribution.

Adapting the C\&W approach to the present setting with per-feature uncertainty constraints, we define the objective
\begin{equation}
\label{eqn:CW}
\mathcal{L}_{\mathrm{hybrid}}
\sum_i \left(\frac{\Delta_i}{\sigma_i}\right)^2
+
c \cdot f!\left(f_\theta(\mathbf{x}^{\mathrm{adv}}), y\right)
+
\lambda_{\chi^2} , \chi^2(\mathbf{x}, \mathbf{x}^{\mathrm{adv}})
+
\lambda_{\mathrm{prior}} , \mathcal{L}_{\mathrm{prior}}(\mathbf{x}, \mathbf{x}^{\mathrm{adv}}),
\end{equation}
where the first term corresponds to a normalized squared perturbation magnitude in units of the feature uncertainties, the second term enforces misclassification, and the last two terms correspond to the distributional and statistical constraints introduced in Eq.~(\ref{eqn:crossattack}). This formulation therefore combines the boundary-focused optimization of the C\&W approach with the physically motivated constraints of the PGD-based construction.

The coefficient $\lambda_{\chi^2}$ penalizes deviations between nominal and adversarial one-dimensional feature distributions, evaluated using differentiable histograms. The coefficient $\lambda_{\mathrm{prior}}$ penalizes non-Gaussian or coherent perturbations through the standardized deviations $z_i=\Delta_i/\sigma_i$, enforcing both small typical deviations and an approximately zero batch mean. 

For binary classification with sigmoid outputs, the margin-based term is defined in terms of the logit as
\begin{equation}
f = \max\bigl(\kappa - s \cdot \mathrm{logit}(f_\theta(\mathbf{x}^{\mathrm{adv}})),, 0\bigr),
\end{equation}
where $s = +1$ for signal and $s = -1$ for background, and $\kappa$ controls the desired confidence of the adversarial misclassification. The condition $f = 0$ corresponds to successful classification with a margin of at least $\kappa$, such that the adversarial example is not merely placed on the decision boundary but pushed beyond it.

In practice, the hybrid attack contains several further hyperparameters that control the balance between adversarial strength and physical realism. The parameter $c$ weights the relative importance of the margin loss and is optimized through a binary search. The confidence parameter $\kappa$ specifies the required margin in logit space; The perturbation range is limited to $N_\sigma$ standard deviations per feature, with the uncertainty scale defined as $\sigma_i=f_i |x_i|$. To avoid vanishing uncertainty scales for features close to zero, we impose a minimum reference scale $|x_i| \rightarrow \max(|x_i|,1)$.

This hybrid formulation differs conceptually from the cross-entropy-based attack in Eq.~(\ref{eqn:crossattack}) in several important aspects. While the PGD approach maximizes the classification loss within a fixed feasible region, it does not explicitly control the size of the perturbation and may drive solutions toward extremal regions of the allowed space. In contrast, the C\&W-inspired objective introduces an explicit penalty on the perturbation magnitude, leading to a continuous trade-off between classification change and minimal deformation. At the same time, the inclusion of the $\chi^2$ and prior terms ensures that the resulting adversarial samples remain statistically consistent with the nominal distribution, preventing the large-scale distortions that would otherwise arise from unconstrained optimization. The margin-based loss operates directly on the decision boundary rather than on probability calibration, avoiding issues associated with saturation of the sigmoid output and providing a more stable optimization landscape.

From the perspective of HEP applications, this combined approach offers several advantages. The perturbation penalty expressed in units of $\sigma_i$ provides a natural, physically interpretable metric that directly quantifies deviations relative to experimental uncertainties. The margin parameter $\kappa$ enables control over the strength of the adversarial effect, allowing the construction of high-confidence perturbations that probe classifier stability beyond marginal boundary crossings. Crucially, the additional distributional constraints ensure that these perturbations remain indistinguishable from nominal samples under standard validation procedures. As a result, the hybrid formulation simultaneously approximates a worst-case deformation within the experimentally allowed region and preserves the statistical realism required for a meaningful interpretation as a potential source of systematic uncertainty.

\section{Benchmark Scenarios and Models}
\label{sec:benchmarks}

We apply this framework to three representative benchmark scenarios based on fast-simulated proton--proton collisions at $\sqrt{s}=13~\mathrm{TeV}$: (i) an event-level DNN classifier separating top-quark pair production from $WW$ events, (ii) a graph neural network–based quark--gluon jet tagger using track-level information, and (iii) a transformer-based classifier for missing transverse energy reconstruction. Across all benchmarks, we observe fooling ratios at the level of $2$--$3\%$ under physically constrained perturbations. Classifiers trained to distinguish nominal from adversarial samples achieve an area under the ROC curve consistent with random guessing when combining classes, indicating statistical indistinguishability at the distributional level. In contrast, simple cut-based approaches exhibit significantly smaller sensitivity. Retraining with adversarial examples reduces, but does not entirely eliminate, the observed effects, leading to performance shifts at the percent level.

These findings suggest that high-capacity classifiers may exhibit a residual model dependence not captured by standard MC systematics or low-dimensional validation checks, motivating dedicated robustness studies as part of the uncertainty-quantification toolkit for future DNN-based HEP analyses.


\subsection{Signal--Background Classification Model}

Signal–background classification is one of the most fundamental and widely used tasks in LHC data analyses. It underpins a broad range of measurements and searches, where the central goal is to distinguish a physics process of interest from often overwhelming backgrounds. Typical examples include the separation of Higgs boson production from Standard Model background processes (e.g.~\cite{ATLAS:2024jry, CMS:2025ihj}), the discrimination of top-quark pair production from electroweak backgrounds (e.g.~\cite{CMS:2025kzt, ATLAS:2024hac, Andrews:2021ejw}), and searches for rare or exotic signals whose signatures resemble well-known processes (e.g.~\cite{CMS:2024zpb, ATLAS:2023ian}). Over time, this task has evolved from simple cut-based selections to multivariate classifiers and, more recently, deep neural networks operating on both high-level and low-level observables.

In this study, we consider a representative event-level classification problem: the separation of top-quark pair production ($t\bar{t}$) from $WW$ production in the semi-leptonic final state. This benchmark captures many essential characteristics of LHC analyses, including moderate feature dimensionality, realistic kinematic correlations, and partially overlapping signal and background phase spaces.

The training samples are generated using \textsc{Pythia8}~\cite{Sjostrand:2007gs} at $\sqrt{s}=13$~TeV and processed through a \textsc{Delphes} ~\cite{deFavereau:2013fsa} fast detector simulation. The detector response is configured to resemble a general-purpose LHC experiment. Both $t\bar{t}$ and $WW$ samples are generated at leading order with standard parton showering and hadronization settings. Events are required to contain one isolated charged lepton and multiple reconstructed jets.

The classifier is implemented as a fully connected deep neural network operating on 12 high-level kinematic observables. These include the transverse momenta and angular variables of the two leading jets, the kinematics of the reconstructed charged lepton, and the total number of reconstructed jets with $p_{\mathrm{T}} > 40$~GeV. This feature set provides a compact yet realistic representation of typical analysis-level inputs used in collider measurements.

\begin{figure}[htbp]
    \centering
    \includegraphics[width=0.32\textwidth]{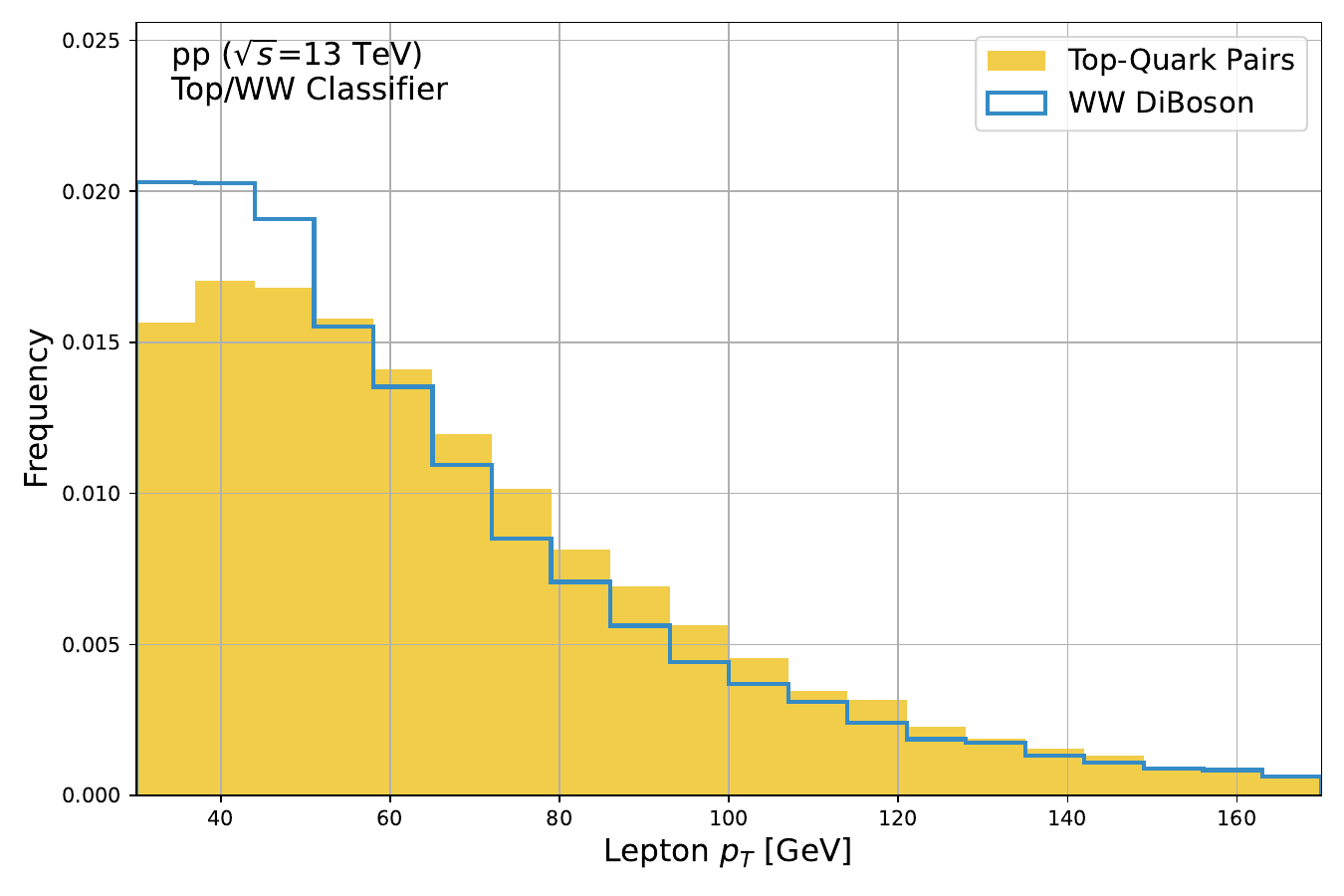}
    \includegraphics[width=0.32\textwidth]{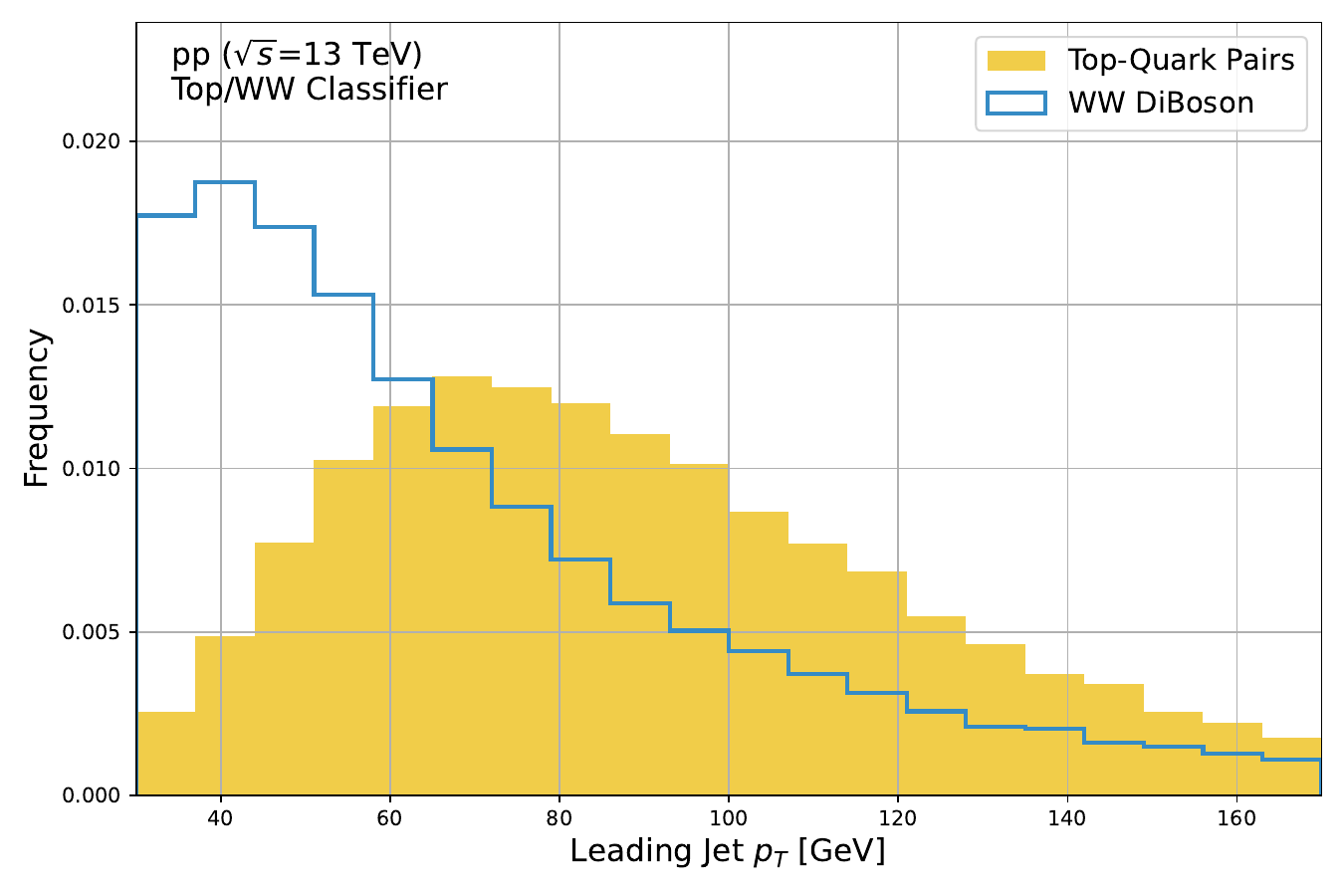}
    \includegraphics[width=0.32\textwidth]{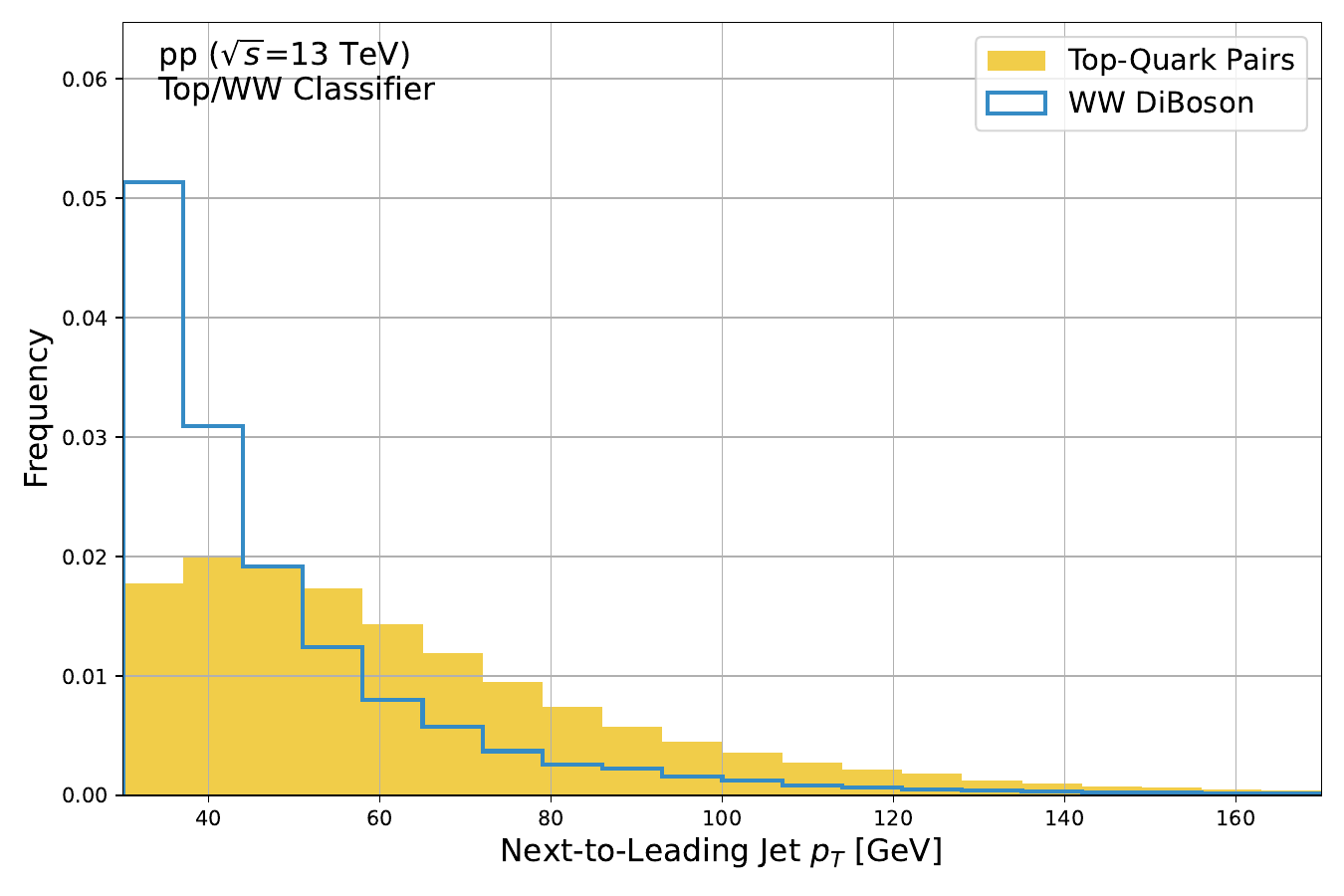}
    \caption{Representative high-level input observables for the $t\bar{t}$ versus $WW$ classification task. Shown are the transverse momentum of the reconstructed lepton (left), the transverse momentum of the leading jet (middle), and the transverse momentum of the subleading jet (right), displayed separately for $t\bar{t}$ and $WW$ events.}
    \label{fig:featuresWWTop}
\end{figure}

Figure~\ref{fig:featuresWWTop} displays three representative high-level input observables used in the $t\bar{t}$ versus $WW$ classification task: the transverse momentum of the reconstructed lepton (left), the transverse momentum of the leading jet (middle), and that of the subleading jet (right). These variables capture key kinematic differences between the two processes while remaining typical of analysis-level inputs in LHC measurements.

Clear but non-trivial differences between $t\bar{t}$ and $WW$ production are visible. In particular, $t\bar{t}$ events tend to exhibit harder jet spectra due to the decay of two heavy top quarks into $b$ quarks and additional hadronic activity. However, no single feature provides sufficient separation power, underscoring the need for multivariate approaches that exploit correlated kinematic information across multiple variables.

In addition to the neural network classifier, we implement a simple cut-based selection as a baseline for comparison. For the $t\bar{t}$ versus $WW$ task, events are classified as $t\bar{t}$ candidates if they contain more than two reconstructed jets, with the leading jet satisfying $p_{\mathrm{T}} > 60$~GeV and the subleading jet $p_{\mathrm{T}} > 40$~GeV. Events failing these requirements are assigned to the $WW$ category. These thresholds reflect the typically higher jet multiplicity and harder jet spectra expected in $t\bar{t}$ production.


\subsection{Quark--Gluon Jet Tagging Model}

Quark–gluon jet tagging is a long-standing and scientifically important problem in collider physics, with direct relevance for both precision measurements and searches for new phenomena at the LHC. The ability to distinguish jets initiated by quarks from those initiated by gluons can significantly improve background rejection in many analyses and plays a particularly important role in precision measurements within the Standard Model. Early approaches relied on hand-crafted observables based on jet shapes and particle multiplicities(e.g. ~\cite{Cornelis:2014ima}). More recently, substantial performance improvements have been achieved using machine-learning techniques that operate directly on low-level jet constituents, including deep neural networks and set-based architectures~(e.g. \cite{Andrews:2019faz}). Prominent examples include energy-flow–based methods and particle-cloud representations, which have established low-level learning as a powerful paradigm for jet tagging~\cite{Qu:2019gqs}.

Graph neural networks (GNNs) provide a particularly natural framework for quark–gluon discrimination, as jets can be interpreted as unordered collections of particles with non-trivial geometric and relational structure in momentum space, e.g. \cite{Gong:2022lye, Semlani:2023kzf}. By explicitly modeling local correlations between nearby particles and aggregating this information into global jet representations, GNNs offer a flexible and physically motivated approach that has demonstrated state-of-the-art performance while remaining robust against variations in detector resolution and particle multiplicity.

We have therefore chosen to formulate the quark--gluon tagging as a graph-learning problem. Jets are reconstructed using the anti-$k_T$ algorithm with radius parameter $R=0.4$. Each jet is represented by up to 50 charged-particle tracks within the jet cone. For each track, the input features include transverse momentum $p_{\mathrm{T}}$, pseudorapidity $\eta$, azimuthal angle $\phi$, electric charge, transverse impact parameter $d_0$, and longitudinal impact parameter $z_0$. Jets with fewer than 50 tracks are zero-padded to maintain a fixed maximum size.

Training samples are derived from $W/Z$+jet events generated with \textsc{Pythia8} at $\sqrt{s}=13$~TeV and processed through \textsc{Delphes} fast detector simulation. To avoid trivial discrimination based solely on kinematic differences, the transverse momentum spectra of quark- and gluon-initiated jets are reweighted to match. The partonic origin of each jet is determined at Monte Carlo truth level by matching the reconstructed jet to the highest-momentum parton within a predefined angular distance.

For each jet, a graph is constructed dynamically using a $k$-nearest-neighbor algorithm in the $(\eta,\phi)$ plane, with proper treatment of the periodicity in $\phi$. Nodes correspond to tracks and edges encode local geometric proximity. The neural network consists of three stacked EdgeConv layers with increasing feature dimensions, each followed by nonlinear activation, batch normalization, and dropout. The outputs of all EdgeConv layers are concatenated and aggregated via global mean pooling to form a fixed-length jet representation, which is subsequently processed by a fully connected classification head producing a single logit.

\begin{figure}[htbp]
    \centering
    \includegraphics[width=0.32\textwidth]{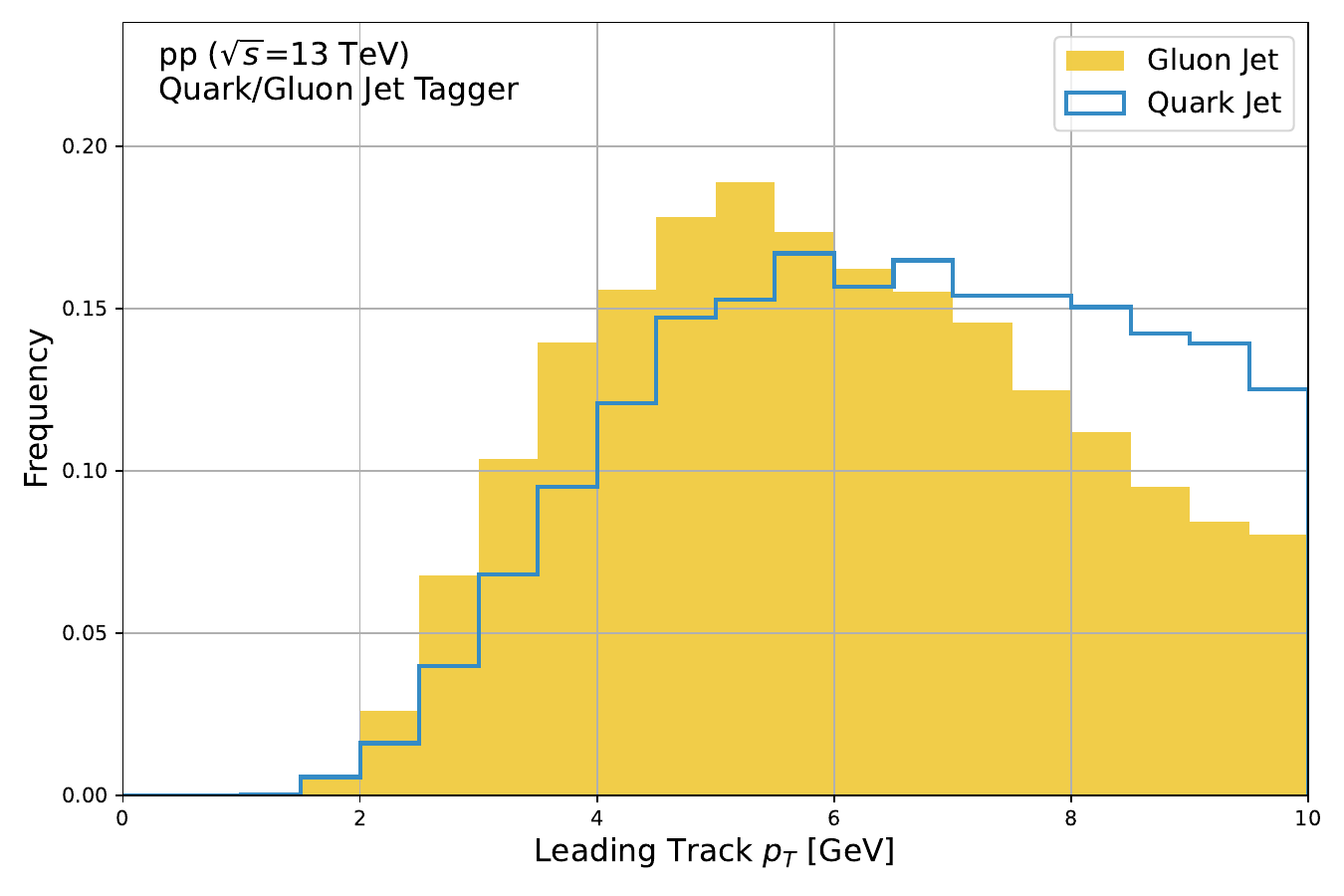}
    \includegraphics[width=0.32\textwidth]{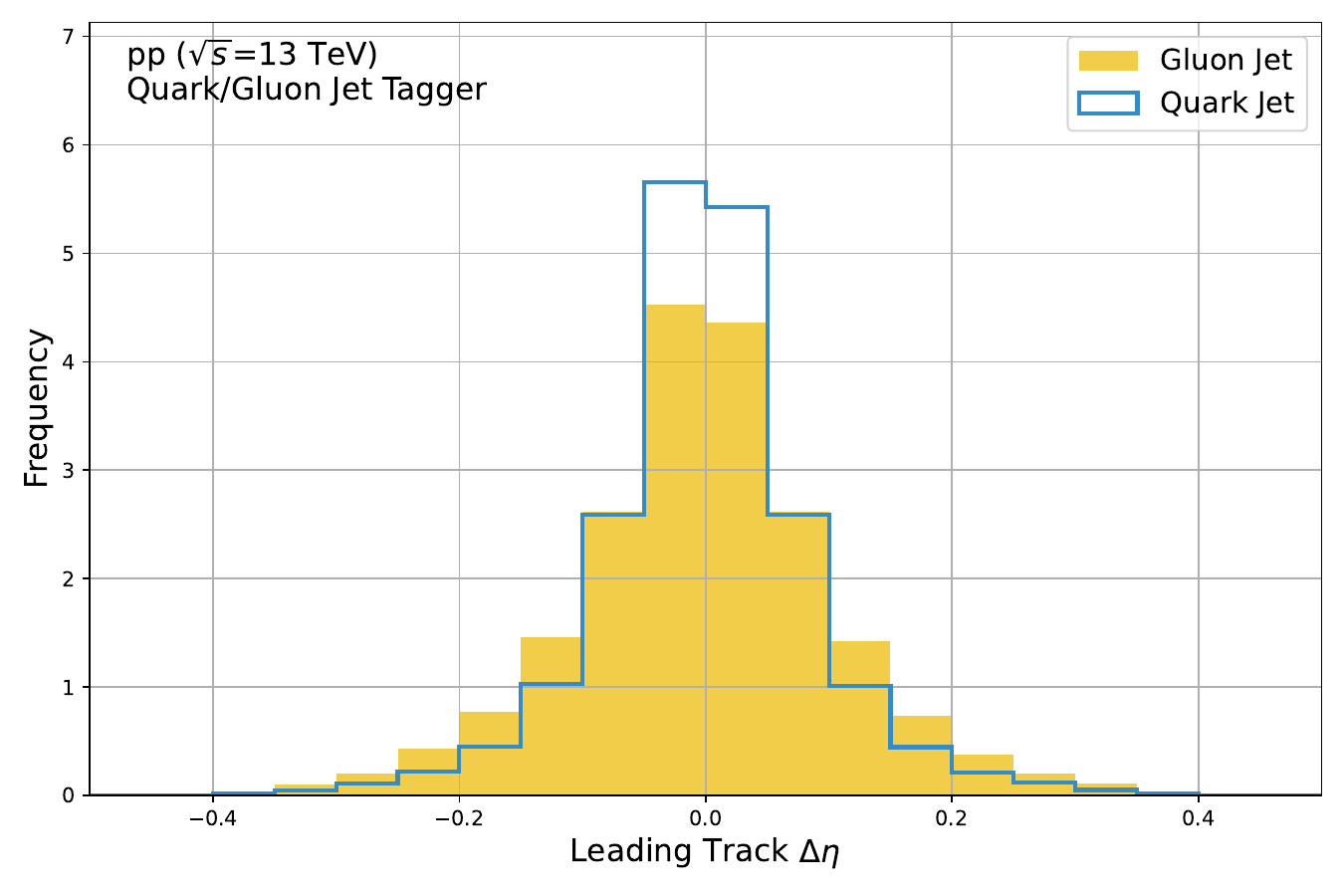}
    \includegraphics[width=0.32\textwidth]{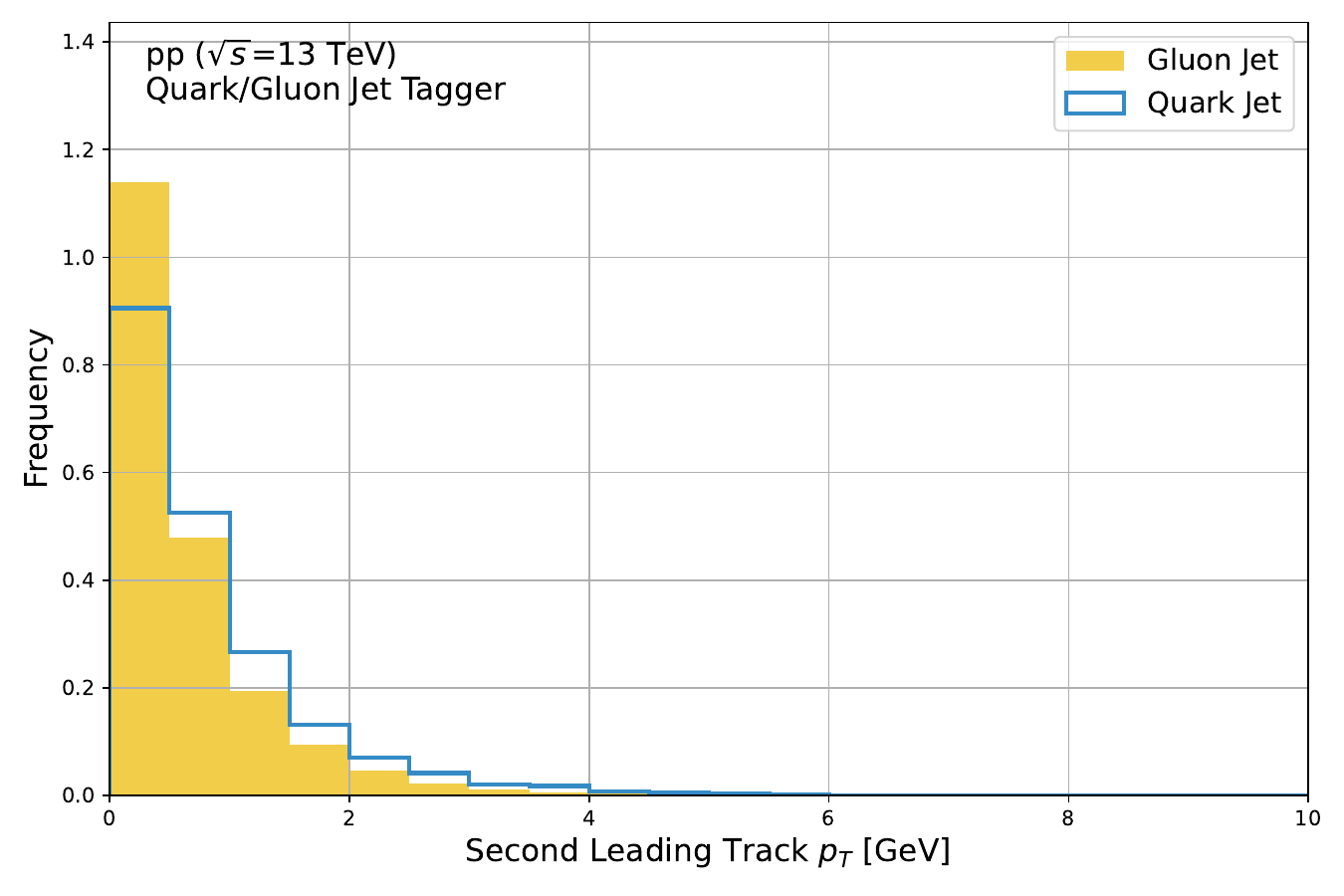}
    \caption{Representative track-level observables for the quark--gluon jet tagging task. Shown are the transverse momentum of the leading track (left), the pseudorapidity difference between the leading track and the jet axis (middle), and the transverse momentum of the second-leading track (right), displayed separately for quark- and gluon-initiated jets.}
    \label{fig:featureJetTagging}
\end{figure}

Figure~\ref{fig:featureJetTagging} presents three representative track-level observables used in the quark--gluon tagging task: the transverse momentum of the leading track (left), the pseudorapidity difference between the leading track and the jet axis (middle), and the transverse momentum of the second-leading track (right). These observables reflect characteristic differences in the radiation patterns of quark- and gluon-initiated jets.  However, substantial overlap between quark and gluon jets remains for all individual observables. This overlap demonstrates that no single feature provides sufficient discriminating power and motivates the use of models capable of learning complex correlations among multiple particles within the jet.

In addition to the graph neural network, we implement a simple cut-based classifier as a baseline reference for the quark--gluon tagging task. For this purpose, each jet is reduced to three commonly used high-level observables derived from its charged-particle constituents: the number of reconstructed tracks $n_{\mathrm{tracks}}$, the momentum dispersion variable $p_{\mathrm{T}}D$, and the jet girth \cite{Larkoski:2014pca}.

The track multiplicity $n_{\mathrm{tracks}}$ reflects the higher particle multiplicity typically observed in gluon jets due to their larger color factor. The variable $p_{\mathrm{T}}^D$ characterizes the dispersion of transverse momentum among jet constituents and is sensitive to the harder fragmentation pattern of quark jets.  The jet girth $g$ quantifies the radial distribution of transverse momentum around the jet axis and is defined as
\begin{equation}
g = \frac{1}{p_{\mathrm{T}}^{\mathrm{jet}}}
\sum_{i \in \mathrm{tracks}} 
p_{\mathrm{T},i} \, \Delta R_i ,
\end{equation}
where $p_{\mathrm{T}}^{\mathrm{jet}} = \sum_i p_{\mathrm{T},i}$ is the total transverse momentum of the jet (restricted here to charged tracks), and
\begin{equation}
\Delta R_i = \sqrt{(\eta_i - \eta_{\mathrm{jet}})^2 + (\phi_i - \phi_{\mathrm{jet}})^2}
\end{equation}
denotes the distance of track $i$ from the jet axis in the $(\eta,\phi)$ plane. Larger values of $g$ correspond to broader jets with radiation distributed at larger angular distances from the jet axis.

The transverse-momentum dispersion variable $p_{\mathrm{T}}D$ characterizes the spread of transverse momentum among jet constituents and is defined as
\begin{equation}
p_{\mathrm{T}}D = 
\frac{\sqrt{\sum_{i \in \mathrm{tracks}} p_{\mathrm{T},i}^2}}
{\sum_{i \in \mathrm{tracks}} p_{\mathrm{T},i}}.
\end{equation}
This observable approaches unity when the jet transverse momentum is dominated by a single hard constituent and decreases as the momentum is distributed more evenly among many softer particles.

Jets are classified as quark candidates if they satisfy the following requirements, $n_{\mathrm{tracks}} \leq 3, \quad
p_{\mathrm{T}}^D < 2.5$, $(0.1 < \mathrm{girth})$ or $(\mathrm{girth} > 0.5)$. Jets failing these criteria are assigned to the gluon category. These thresholds are chosen to reflect the characteristic lower multiplicity and more collimated structure of quark jets compared to gluon jets. While such a cut-based approach captures the dominant physical trends, it does not exploit detailed multi-particle correlations and therefore provides a natural reference against which the performance and robustness of the graph-based model can be evaluated.


\subsection{Missing Transverse Energy Reconstruction Model}

Missing transverse energy ($E_{\mathrm{T}}^{\mathrm{miss}}$) is a key observable in many LHC analyses and is particularly sensitive to detector effects, soft radiation, and pile-up contributions. Its reconstruction involves the vectorial combination of multiple reconstructed objects and is therefore intrinsically a high-dimensional problem. Traditional approaches rely on calibrated sums of reconstructed particles and objects, but their performance can be limited by detector resolution and complex event environments. Recently, machine-learning techniques have been explored to improve the reconstruction and resolution of $E_{\mathrm{T}}^{\mathrm{miss}}$ by leveraging low-level detector information and learning correlations among reconstructed particles~\cite{CMS:2025prt}. In particular, attention-based architectures such as transformer models have emerged as a powerful paradigm for processing sets or sequences of particles, enabling the learning of global event-level correlations through self-attention mechanisms~\cite{Qu:2022mxj, 10.21468/SciPostPhysCore.8.1.026}. These architectures naturally handle variable-length inputs and have shown strong performance in a variety of particle-physics tasks, making them well suited for complex event-level reconstruction problems such as $E_{\mathrm{T}}^{\mathrm{miss}}$.

In this work, the task is formulated as a binary classification problem using an transformer-based architecture: events are classified according to whether $E_{\mathrm{T}}^{\mathrm{miss}} > 60$~GeV or $E_{\mathrm{T}}^{\mathrm{miss}} < 60$~GeV. This choice simplifies the definition of adversarial perturbations while retaining the essential complexity of the reconstruction problem.

The training samples are based on $W \rightarrow \mu \nu$ events generated with \textsc{Pythia8} \cite{Sjostrand:2007gs} at $\sqrt{s}=13$~TeV and processed through \textsc{Delphes} detector simulation \cite{deFavereau:2013fsa}. Each event is represented by its reconstructed charged-particle tracks. For each track, the input features are the momentum components $(p_x, p_y, p_z)$ and the transverse impact parameter $d_0$. The number of tracks varies event by event. The transformer-based architecture is employed to process the variable-length sequence of tracks. Each track is first embedded into a fixed-dimensional latent space via a linear projection followed by a GELU activation and layer normalization. A learnable classification token is prepended to the sequence to aggregate global event information. The embedded sequence is then processed by multiple transformer encoder layers with multi-head self-attention and feed-forward sublayers. The output corresponding to the classification token is passed through a fully connected head that produces a single logit representing the $E_{\mathrm{T}}^{\mathrm{miss}}$ class.

\begin{figure}[htbp]
    \centering
    \includegraphics[width=0.32\textwidth]{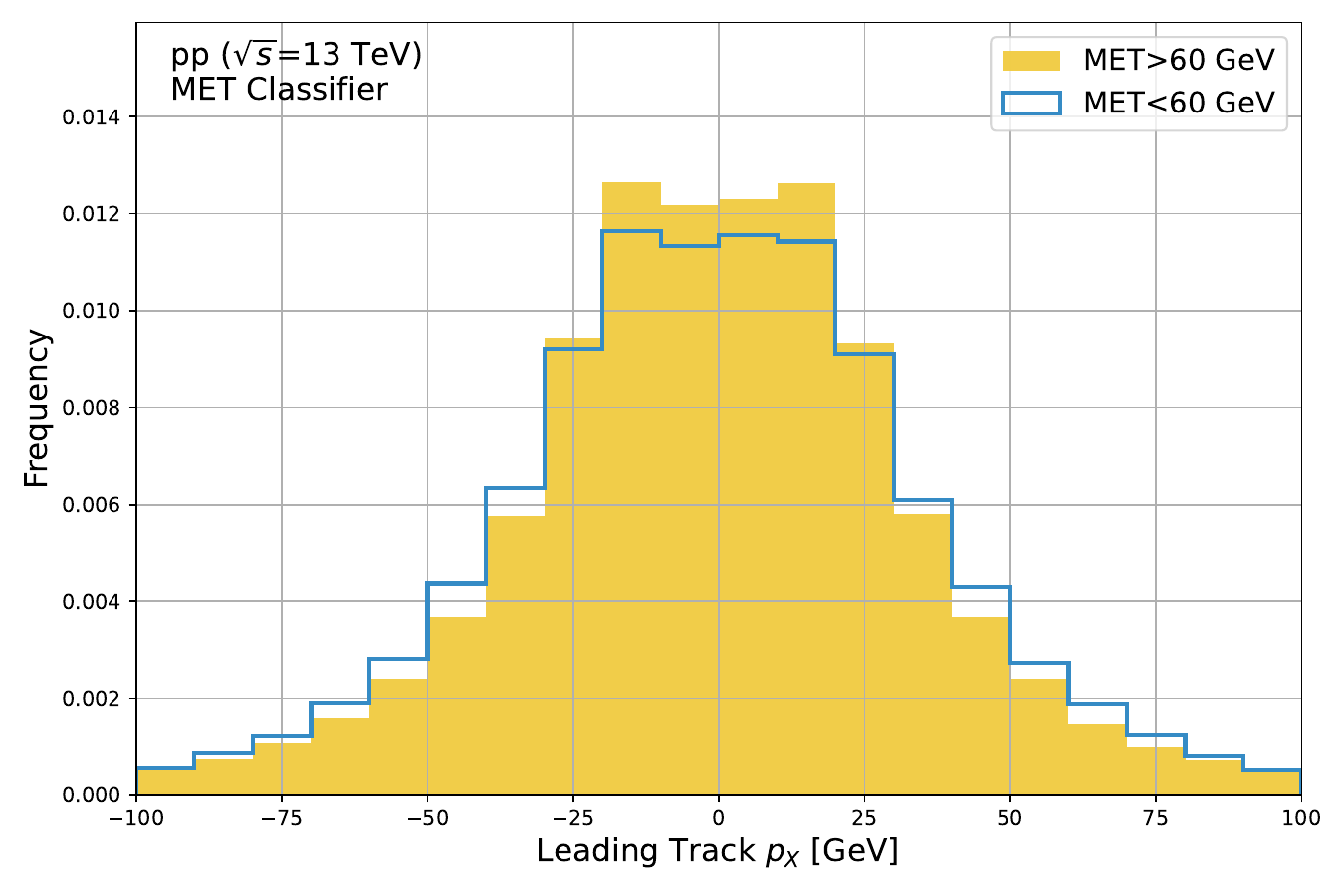}
    \includegraphics[width=0.32\textwidth]{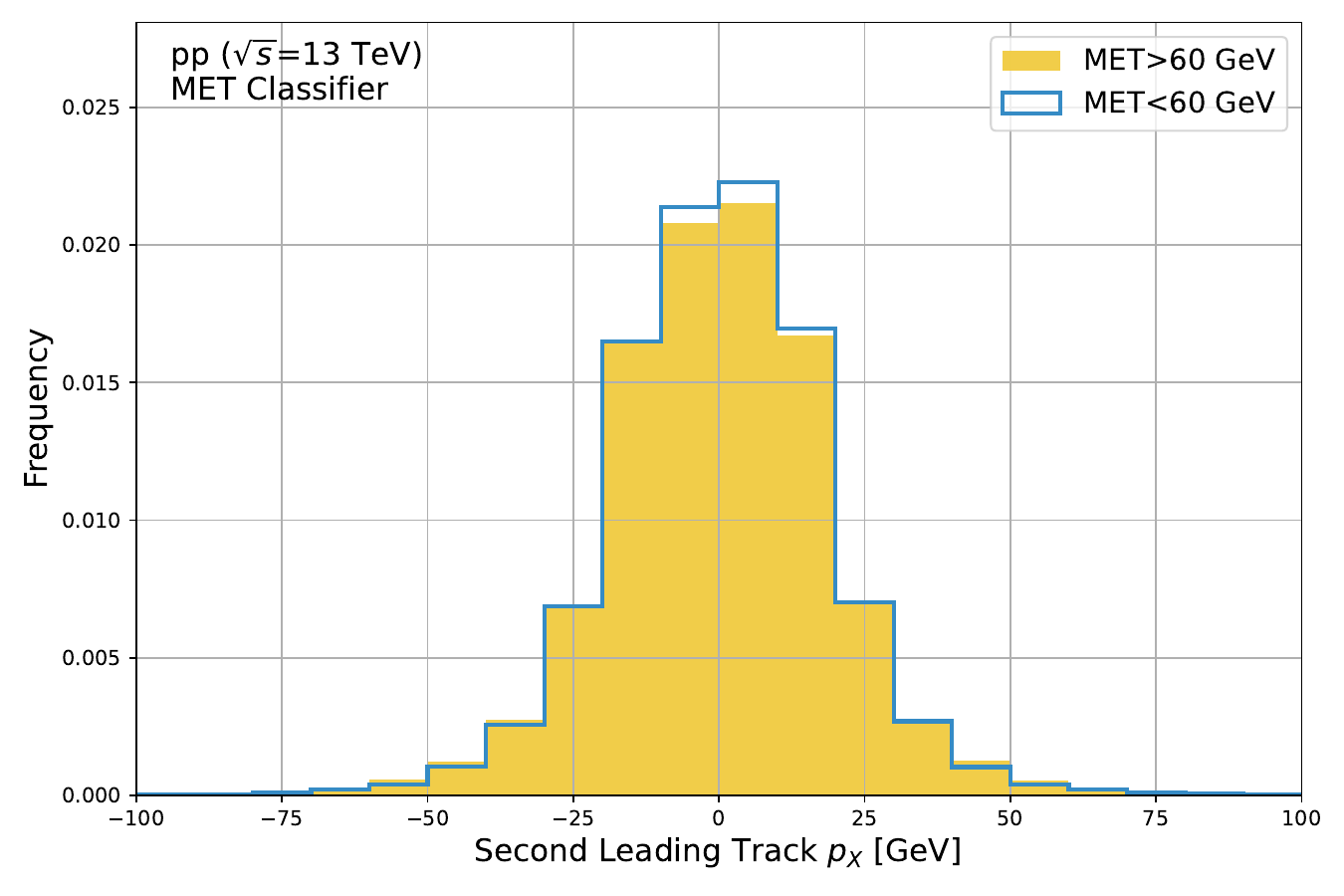}
    \includegraphics[width=0.32\textwidth]{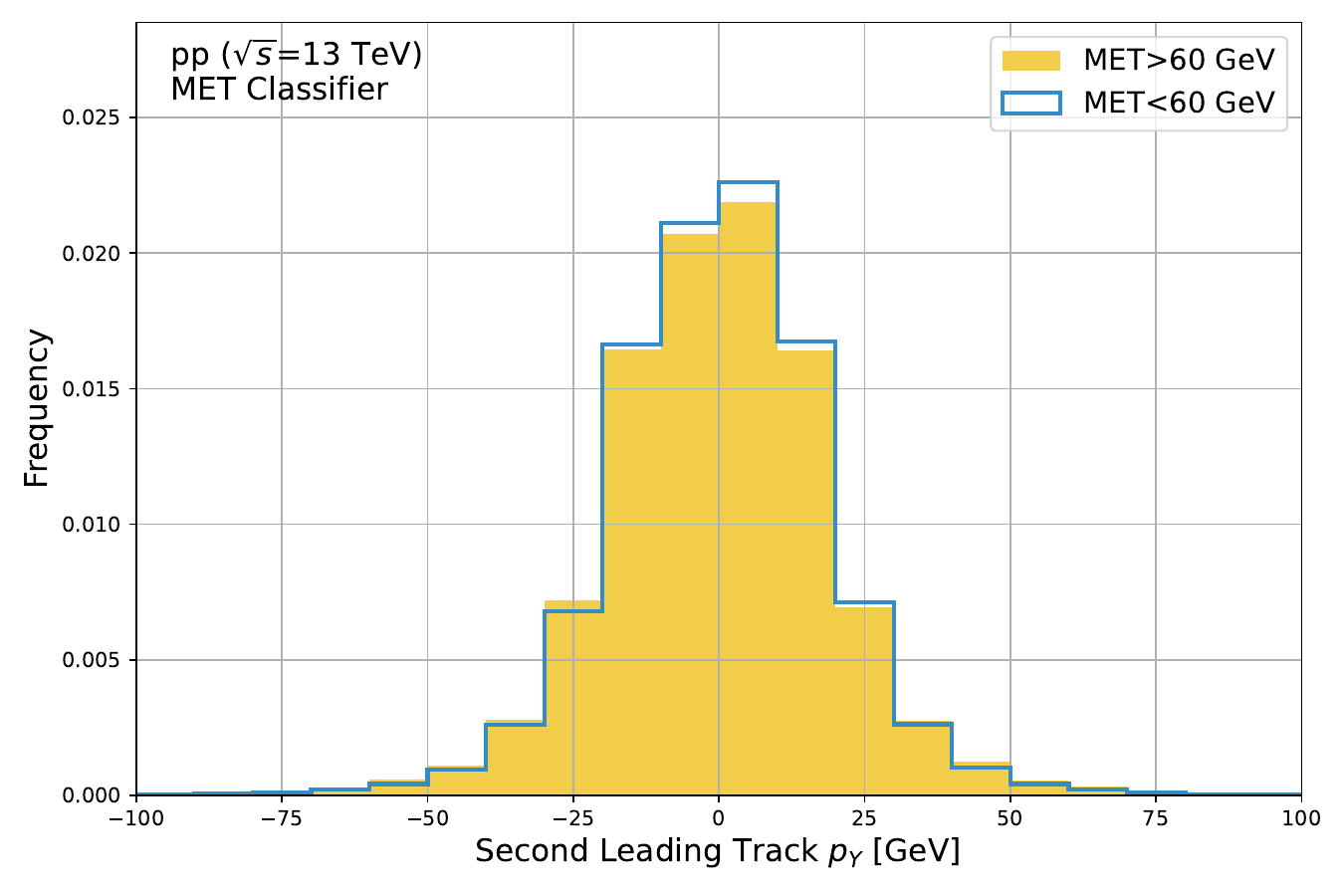}
    \caption{Representative track-level input features for the $E_{\mathrm{T}}^{\mathrm{miss}}$ classification task. Shown are the $p_x$ component of the leading track (left), and the $p_x$ (middle) and $p_y$ (right) components of the second-leading track. Events are separated according to $E_{\mathrm{T}}^{\mathrm{miss}} > 60$~GeV and $E_{\mathrm{T}}^{\mathrm{miss}} < 60$~GeV.}
    \label{fig:featureETMiss}
\end{figure}

Figure~\ref{fig:featureETMiss} shows representative input feature distributions for the $E_{\mathrm{T}}^{\mathrm{miss}}$ classification task. Displayed are the $p_x$ component of the leading track (left), as well as the $p_x$ (middle) and $p_y$ (right) components of the second-leading track. In $W \rightarrow \mu \nu$ events, the leading track typically corresponds to the high-$p_{\mathrm{T}}$ muon, resulting in a pronounced high-momentum tail in the $p_x$ distribution. The second-leading track, often associated with hadronic activity or underlying event contributions, exhibits a significantly softer spectrum. The distributions are shown separately for events with $E_{\mathrm{T}}^{\mathrm{miss}} > 60$~GeV and $E_{\mathrm{T}}^{\mathrm{miss}} < 60$~GeV. 

In addition to the transformer-based model, we define a simple cut-based baseline for the $E_{\mathrm{T}}^{\mathrm{miss}}$ classification task. In this approach, the missing transverse momentum vector is reconstructed by directly summing the transverse momentum components of all reconstructed charged-particle tracks in the event,
\begin{equation}
\vec{p}_{\mathrm{T}}^{\,\mathrm{miss}} = - \sum_{i}^{\mathrm{tracks}} \left( p_{x,i},\, p_{y,i} \right),
\end{equation}
and computing its magnitude, $E_{\mathrm{T}}^{\mathrm{miss}} = \left| \vec{p}_{\mathrm{T}}^{\,\mathrm{miss}} \right|$. No additional quality criteria or impact-parameter requirements are applied, and the transverse impact parameter $d_0$ is ignored. Consequently, this baseline includes contributions from pile-up and detector effects without mitigation, reflecting a minimal reconstruction strategy based solely on track momenta. Events are then classified according to whether the reconstructed $E_{\mathrm{T}}^{\mathrm{miss}}$ exceeds 60~GeV or not. While this approach captures the dominant transverse momentum imbalance in the event, it does not exploit correlations among tracks or attempt to suppress pile-up contributions.

\section{Evaluation, Results and Discussion}
\label{sec:results}

\subsection{Evaluation Strategy}

For each of the three benchmark models, we perform the following studies in a structured sequence designed to isolate nominal performance, quantify adversarial sensitivity, and assess robustness under retraining:

\begin{itemize}

\item \textbf{Nominal training and baseline performance determination: }  
The available dataset is split into 80\% training, 10\% validation, and 10\% test samples. 
Training is performed using early stopping based on the validation loss to prevent overfitting. 
The final model performance is evaluated exclusively on the independent test set.  
This establishes the nominal reference performance under standard training conditions.

\item \textbf{Generation of adversarial samples: }  
Adversarial examples are constructed for the full dataset using the uncertainty-constrained attack described above. 
The adversarial samples are subsequently partitioned into training, validation, and test subsets following the same split as the original data. By construction, each training example has a corresponding adversarial counterpart, and the two are therefore not statistically independent. However, since classifiers are trained exclusively on a given training set (either nominal or adversarial) and evaluated on the corresponding test set, this dependence does not bias the performance comparison. 

\item \textbf{Indistinguishability validation: }  
To verify that the adversarial samples are statistically indistinguishable from nominal events, a secondary classifier with identical architecture is trained to separate nominal and adversarial examples. 
In all cases, the resulting area under the ROC curve (AUC) is found to be close to 0.5, confirming that the adversarial perturbations cannot be identified by a dedicated classifier.  
This step is essential to demonstrate that the attack does not introduce trivial or detectable distortions.

\item \textbf{Statistical stability assessment: }  
All trainings are repeated multiple times with different random initializations and dataset partitions. 
Mean performance values and their root-mean-square (RMS) spread are reported.  
This procedure quantifies statistical fluctuations and ensures that observed effects are robust against training instabilities.

\item \textbf{Nominal performance metrics: }  
Classifier performance is quantified using the signal selection efficiency at a fixed background rejection rate, as well as the ROC AUC. 
The reference background rejection working point is determined from the corresponding cut-based baseline classifier.  
This provides a physics-motivated and model-independent comparison point.

\item \textbf{Performance degradation under adversarial perturbations: }  
The nominally trained classifier is first evaluated on the nominal test set and subsequently on the adversarial test set. 
The difference in performance directly quantifies the sensitivity of the model to physically allowed perturbations and serves as a measure of intrinsic model dependence.

\item \textbf{Assessment of robustness and potential mitigation strategies:} To assess robustness and potential mitigation strategies, two additional training configurations are considered: First, we conduct a training exclusively on adversarial samples and testing on the nominal test set. Second, we train in addition on a combined dataset of nominal and adversarial samples and testing on the nominal test set. Comparing these results to the nominal baseline performance allows us to evaluate whether adversarial exposure stabilizes the classifier and to what extent performance shifts persist.
\end{itemize}

\subsection{DNN Based Classifiers}

As a representative example of a DNN-based classifier, we consider the $t\bar{t}$ versus $WW$ tagger described above. The model is implemented as a fully connected multilayer perceptron with three hidden layers of sizes 64, 64, and 32, ReLU activations, and dropout regularization. In total, the network comprises approximately 7\,000 trainable parameters. Despite its relatively modest size, this architecture already captures non-trivial correlations among the 12 high-level kinematic input features and significantly outperforms the cut-based baseline. In fact, architectures of comparable complexity are widely used in published and ongoing LHC analyses.

The $t\bar{t}$ versus $WW$ classification study is performed on a dataset of 100\,000 simulated events. Adversarial examples are generated using the uncertainty-constrained projected gradient descent algorithm (see chapter \ref{sec:framework_attack}) with a step size $\alpha = 0.04$ and 20 optimization steps. The maximal per-feature deviation is limited to $3\sigma$, and the regularization strengths are set to $\lambda_{\Delta}=0.5$ and $\lambda_{\chi^2}=0.5$. Per-feature uncertainties are defined relative to the observable magnitude via $\sigma_i = f_i |x_i|$, with fractional widths of 2\% for transverse momentum, 0.1\% for pseudorapidity and azimuthal angle, 0.5\% for the impact parameters $d_0$ and $z_0$, and no variation applied to the electric charge. 

\begin{figure}[htbp]
    \centering
    \includegraphics[width=0.32\textwidth]{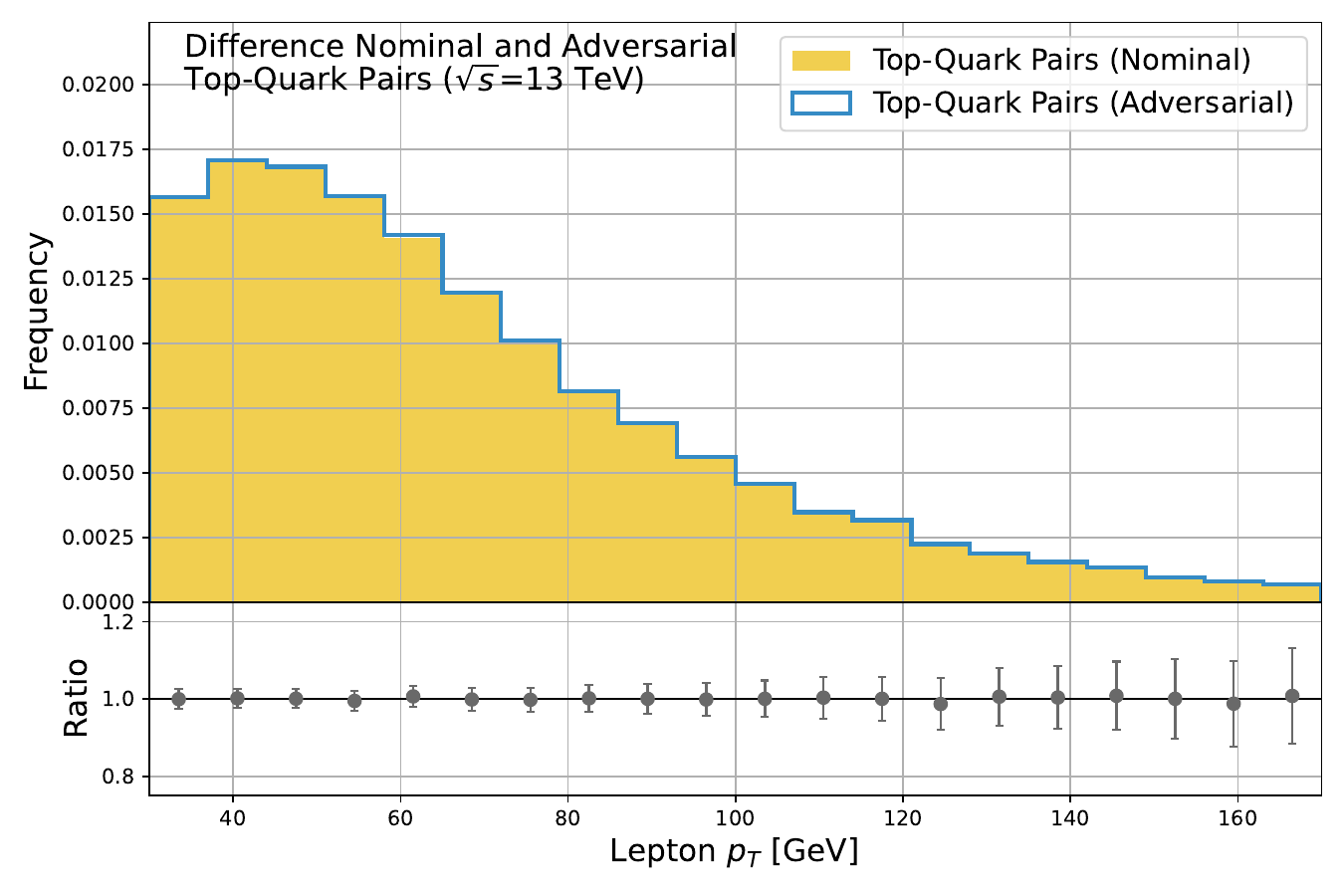}
    \includegraphics[width=0.32\textwidth]{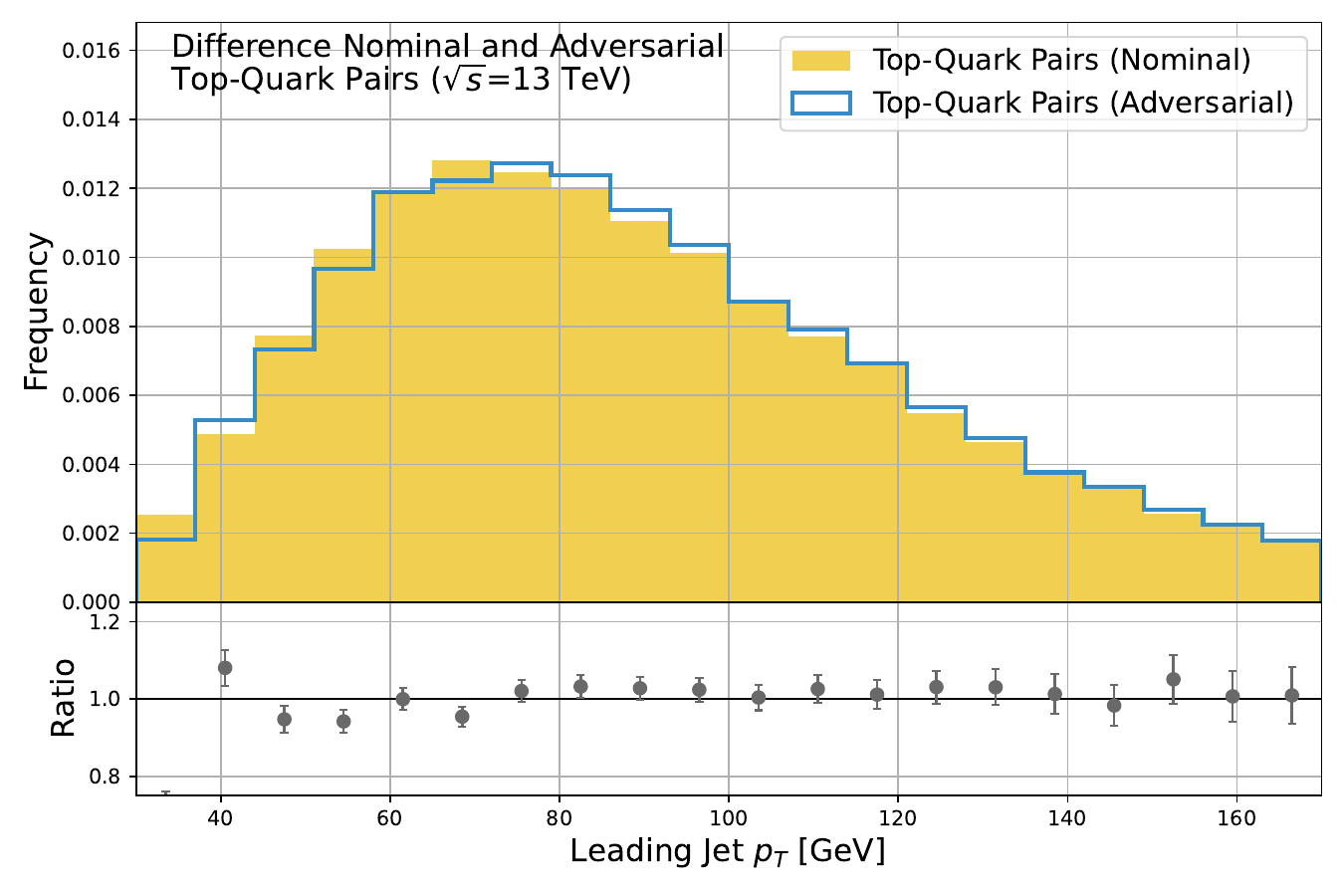}
    \includegraphics[width=0.32\textwidth]{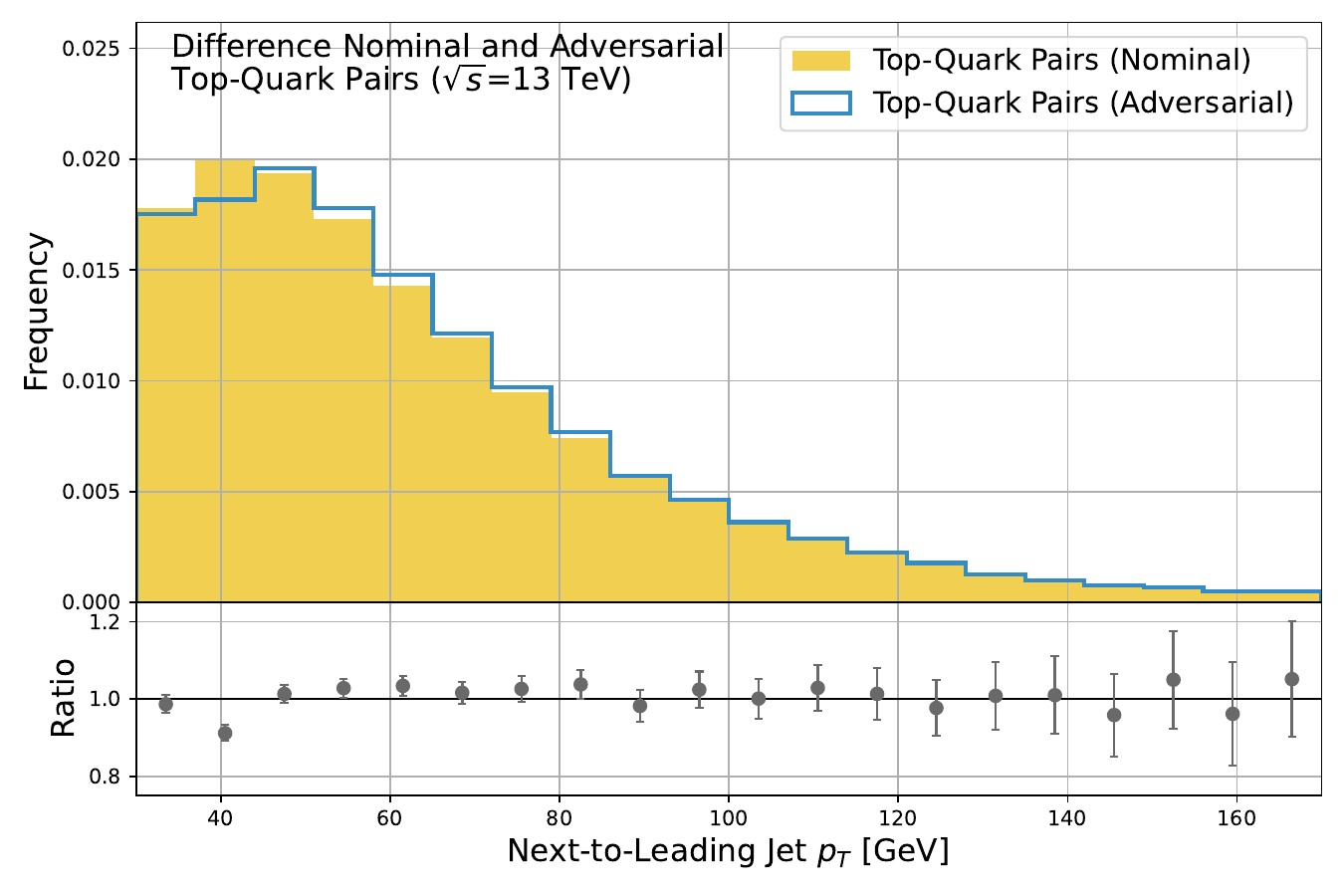}

    \includegraphics[width=0.32\textwidth]{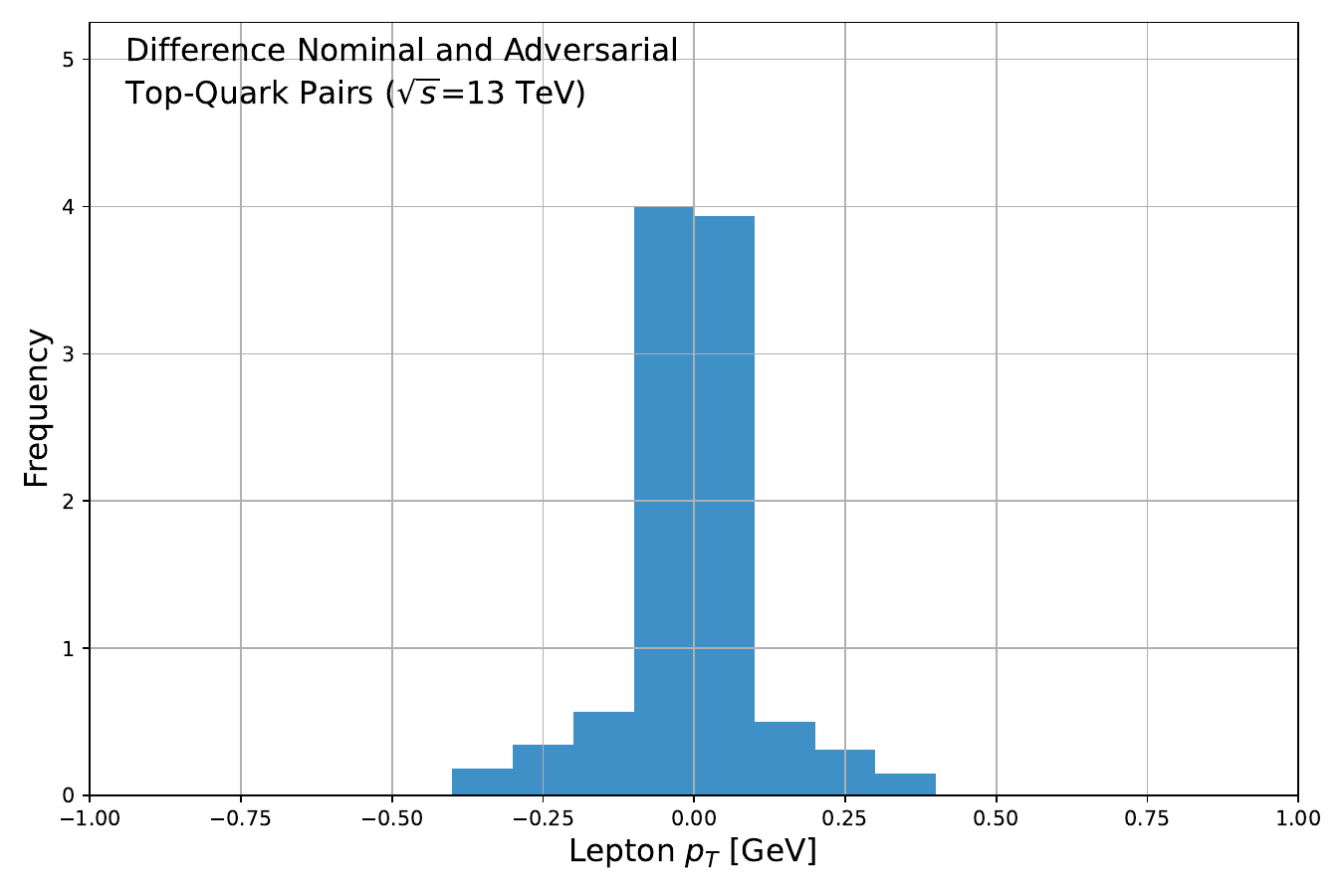}
    \includegraphics[width=0.32\textwidth]{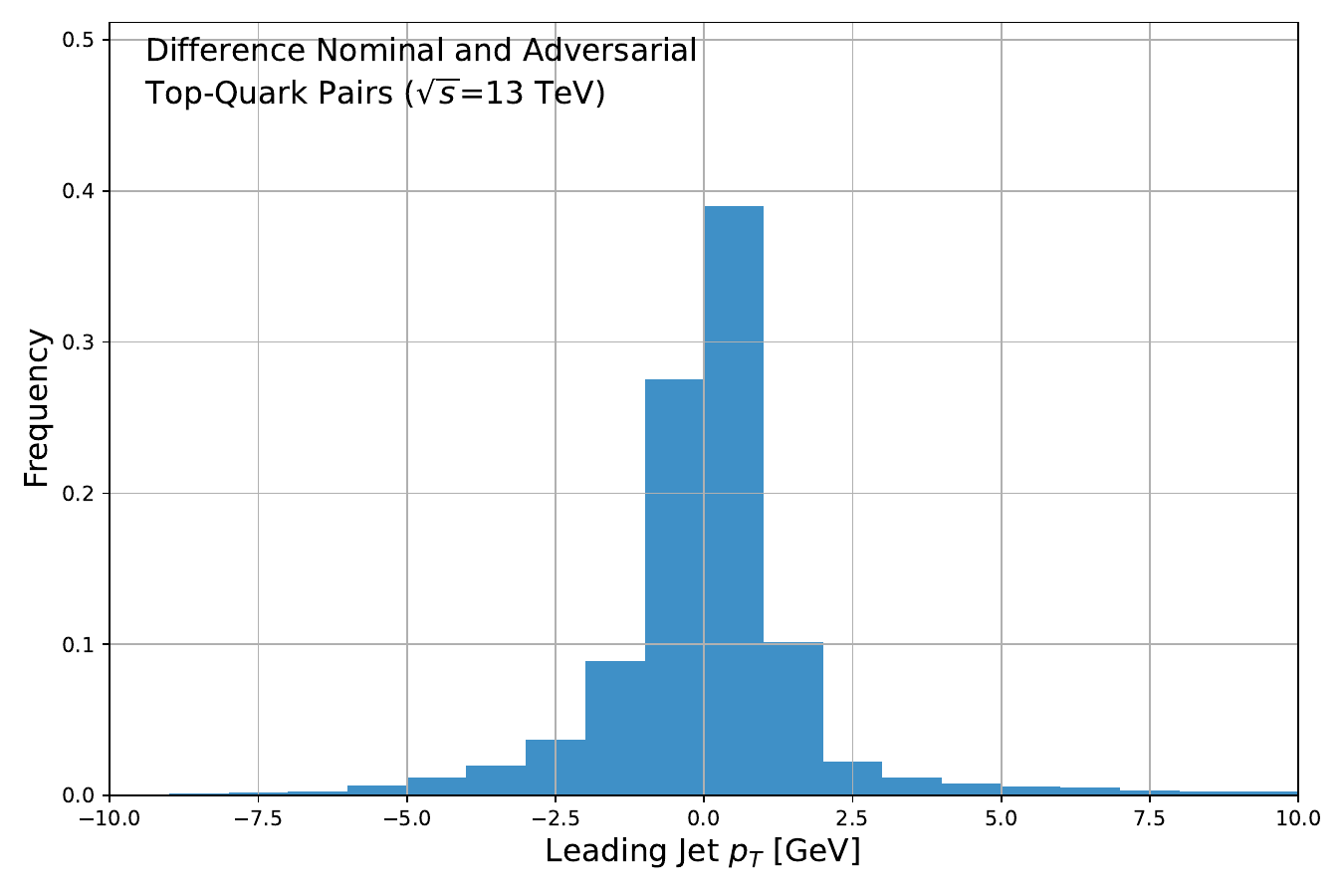}
    \includegraphics[width=0.32\textwidth]{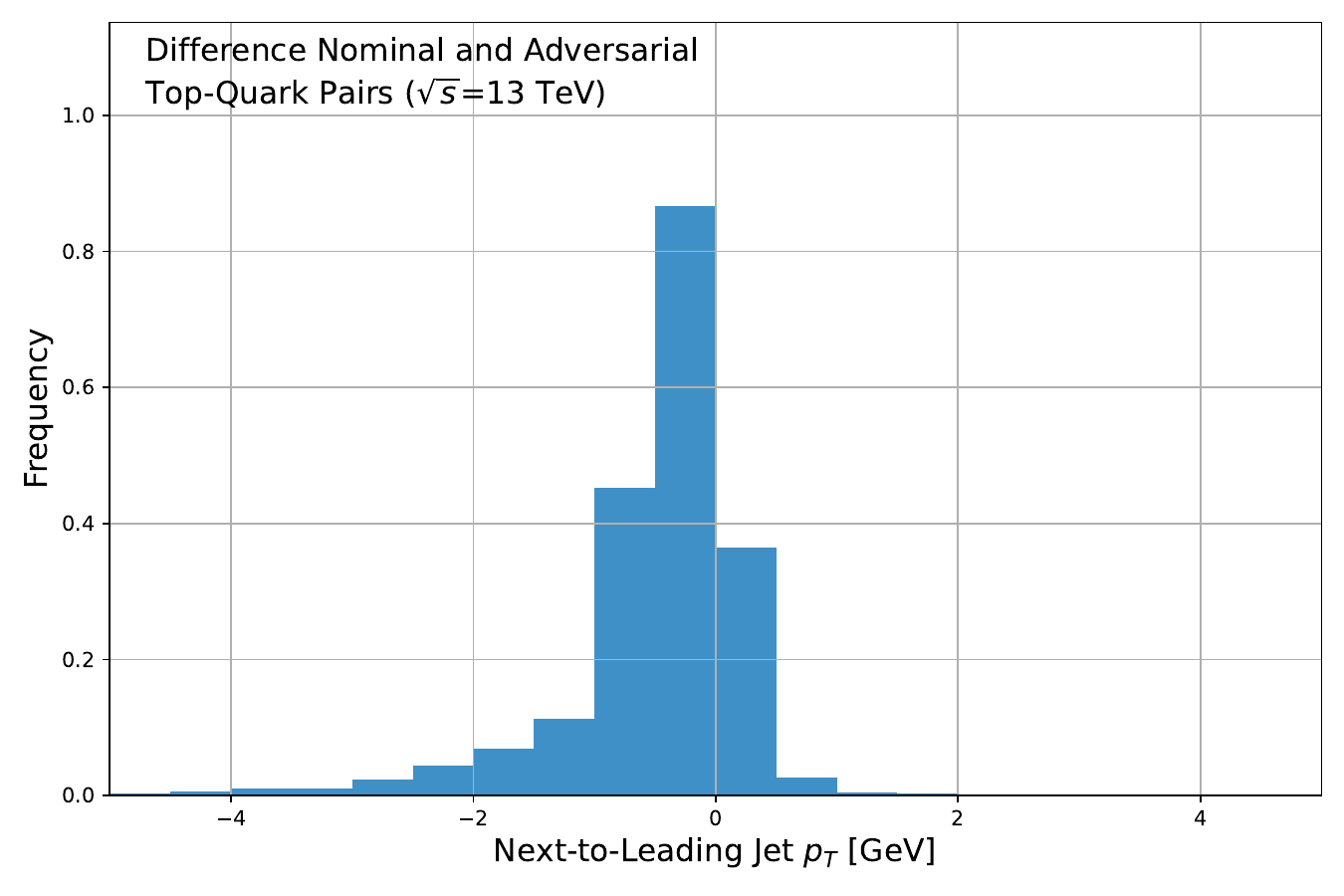}
\caption{Results for the $t\bar{t}$ versus $WW$ classifier. 
Top row: Comparison of the nominal (yellow) and adversarial (blue) distributions for three representative input observables in $t\bar{t}$ events: the transverse momentum of the reconstructed lepton (left), the transverse momentum of the leading jet (middle), and the transverse momentum of the subleading jet (right). Bottom row: event-by-event differences between the nominal and adversarial distributions for the same observables. \label{fig:adverWWTop}}
\end{figure}

\begin{figure}[htbp]
    \centering
    \includegraphics[width=0.49\textwidth]{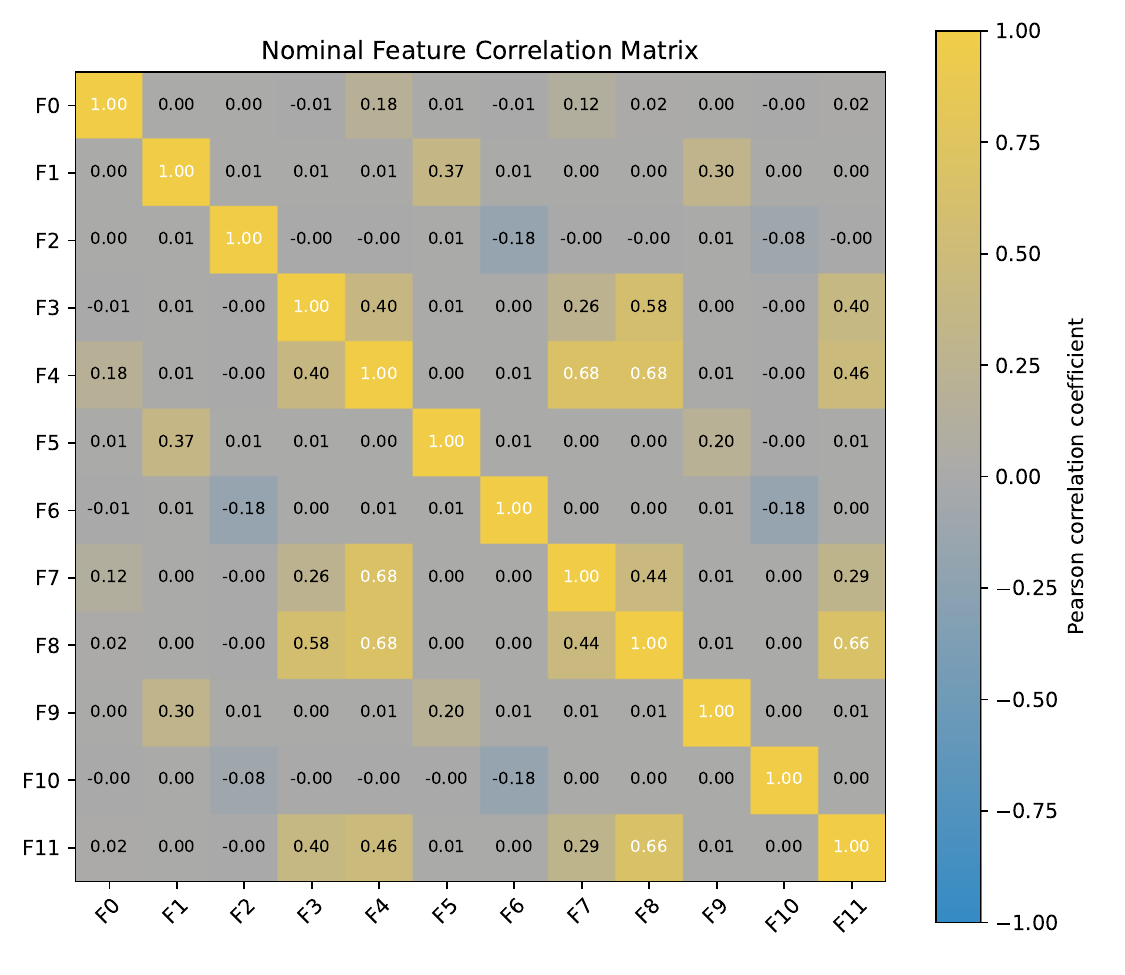}
    \includegraphics[width=0.49\textwidth]{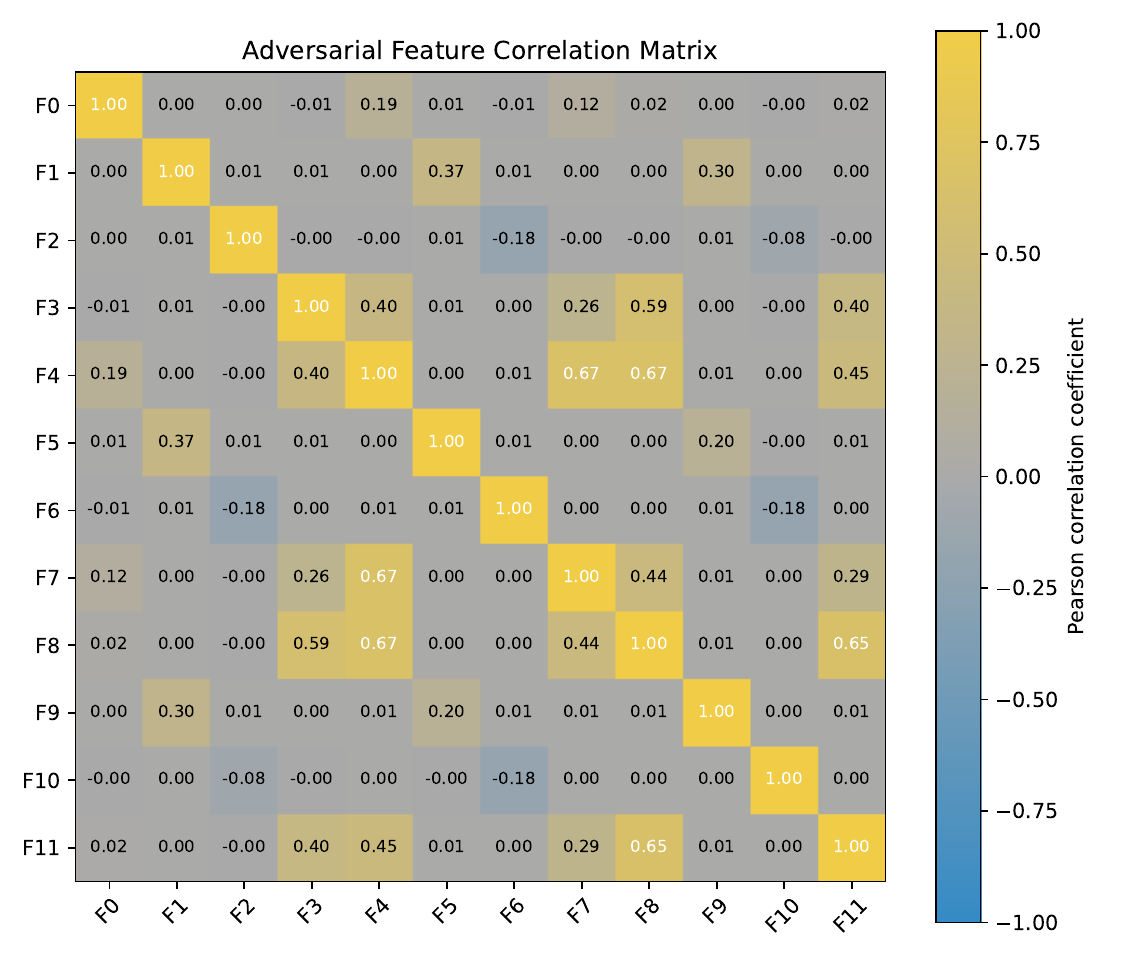}
    \caption{Correlation between the input features for the $t\bar{t}$ versus $WW$ classifier for the nominal (left) and the adversarial samples (right). \label{fig:WWTTBarCorrelation}}
\end{figure}

The adversarial results for the $t\bar{t}$ versus $WW$ classifier are shown in Figure~\ref{fig:adverWWTop}, where the top row compares the nominal (yellow) and adversarial (blue) distributions of three representative input observables in $t\bar{t}$ events, and the bottom row displays the corresponding bin-by-bin differences between the two distributions. The Pearson correlation coefficients between the input features remain largely unchanged  when comparing nominal and adversarial samples, as illustrated in Figure~\ref{fig:WWTTBarCorrelation}. As described, a classifier with identical architecture was also trained to distinguish nominal signal events from their adversarial counterparts, yielding an area under the ROC curve of $0.495 \pm 0.002$. 

The results are summarized in Table~\ref{tab:adversarial_comparison_testset_variant}. 
The cut-based baseline shows no sensitivity to the adversarial perturbations, with identical signal efficiency and background rejection for nominal and adversarial evaluation samples within statistical uncertainties. It should be noted that the given statistical uncertainties between the nominal and adversarial evaluation samples are highly correlated as they are based on the same initial dataset. 

In contrast, the DNN trained exclusively on nominal data exhibits a measurable performance degradation when evaluated on adversarial events. The signal efficiency at fixed background rejection decreases by $0.020 \pm 0.005$, and the ROC AUC is reduced by $0.027 \pm 0.001$. This demonstrates that the classifier decision boundary is sensitive to perturbations that remain invisible in standard validation distributions.

\begin{table}[htbp]
\centering
\caption{Performance comparison for the WW/$t\bar t$ classifier between nominal and adversarial evaluation samples on the test set for different training strategies.}
\label{tab:adversarial_comparison_testset_variant}
\begin{tabular}{l|ccc}
\hline
 & Nominal Eval. & Adversarial Eval. & Difference \\
\hline
\multicolumn{4}{l}{\textbf{Baseline Performance}} \\
\hline
Signal Efficiency & $0.394 \pm 0.003$ & $0.394 \pm 0.003$ & $0.000 \pm 0.004$ \\
Background Rejection & $0.876 \pm 0.002$ & $0.876 \pm 0.002$ & $0.000 \pm 0.002$ \\
\hline
\multicolumn{4}{l}{\textbf{DNN -- Trained on Nominal Sample Only}} \\
\hline
Signal Efficiency (at fixed background rejection) & $0.600 \pm 0.004$ & $0.580 \pm 0.003$ & $0.020 \pm 0.005$ \\
ROC AUC & $0.842 \pm 0.001$ & $0.815 \pm 0.001$ & $0.027 \pm 0.001$ \\
\hline
\multicolumn{4}{l}{\textbf{DNN -- Trained on Adversarial Sample Only }} \\
\hline
Signal Efficiency (at fixed background rejection) & $0.580 \pm 0.006$ & $0.588 \pm 0.004$ & $-0.008 \pm 0.007$ \\
ROC AUC & $0.824 \pm 0.003$ & $0.834 \pm 0.008$ & $-0.011 \pm 0.008$ \\
\hline
\multicolumn{4}{l}{\textbf{DNN -- Trained on Nominal + Adversarial Sample }} \\
\hline
Signal Efficiency (at fixed background rejection) & $0.603 \pm 0.005$ & $0.614 \pm 0.004$ & $-0.010 \pm 0.006$ \\
ROC AUC & $0.840 \pm 0.001$ & $0.846 \pm 0.006$ & $-0.006 \pm 0.006$ \\
\hline
\hline
\multicolumn{4}{l}{\textbf{DNN -- Trained on Nominal Sample Only (Alternative C\&W Loss Function)}} \\
\hline
Signal Efficiency (at fixed background rejection) & $0.600 \pm 0.004$ & $0.572 \pm 0.004$ & $0.028 \pm 0.005$ \\
ROC AUC & $0.842 \pm 0.001$ & $0.822 \pm 0.001$ & $0.02 \pm 0.001$ \\
\hline
\end{tabular}
\end{table}

When the DNN is trained solely on adversarial samples, its performance on nominal data is already reduced compared to the nominal baseline, indicating that the adversarial deformation modifies the effective feature space learned by the network. However, the difference between nominal and adversarial evaluation becomes small within uncertainties, suggesting partial adaptation to the perturbed distribution.

Finally, training on the combined nominal and adversarial dataset largely restores the nominal performance while simultaneously stabilizing the classifier against adversarial evaluation. The remaining differences between nominal and adversarial test samples are small and statistically compatible with zero within uncertainties. 

Overall, these results indicate that high-capacity classifiers exhibit a non-negligible sensitivity to physically allowed perturbations, while cut-based methods remain essentially unaffected. At the same time, exposure to adversarial examples during training can mitigate, though not entirely eliminate, this intrinsic model dependence.

\begin{figure}[htbp]
    \centering
    \includegraphics[width=0.32\textwidth]{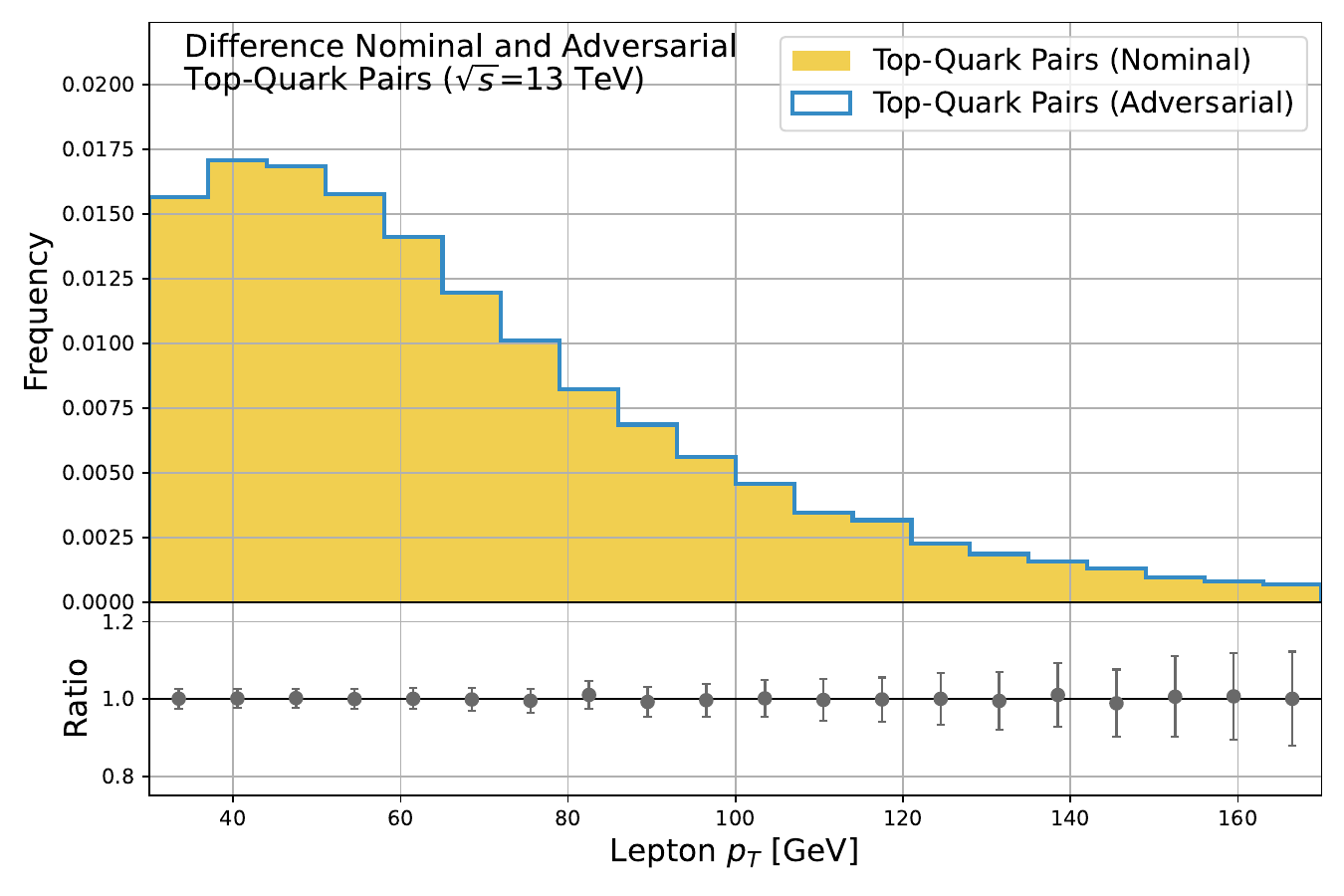}
    \includegraphics[width=0.32\textwidth]{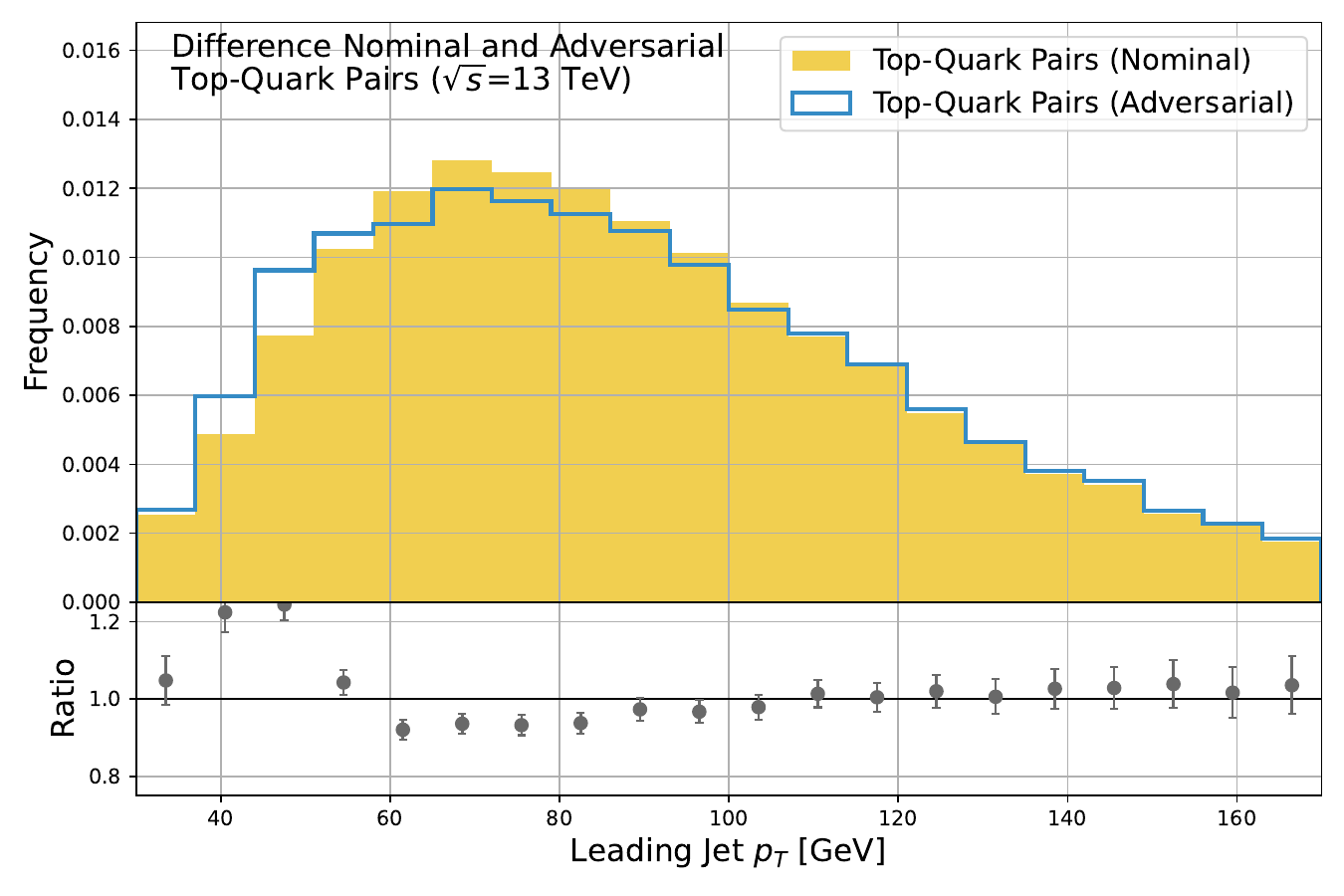}
    \includegraphics[width=0.32\textwidth]{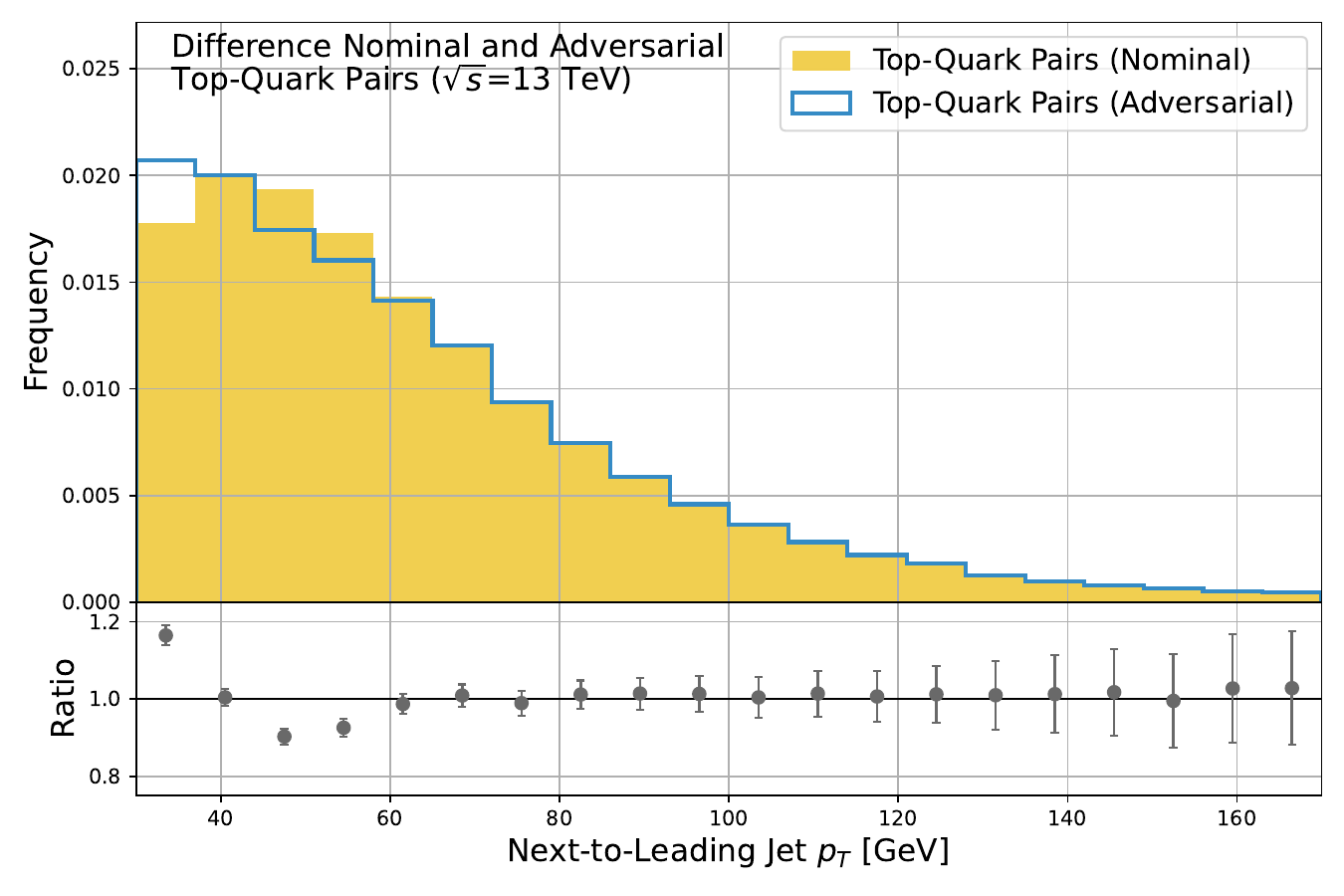}

    \includegraphics[width=0.32\textwidth]{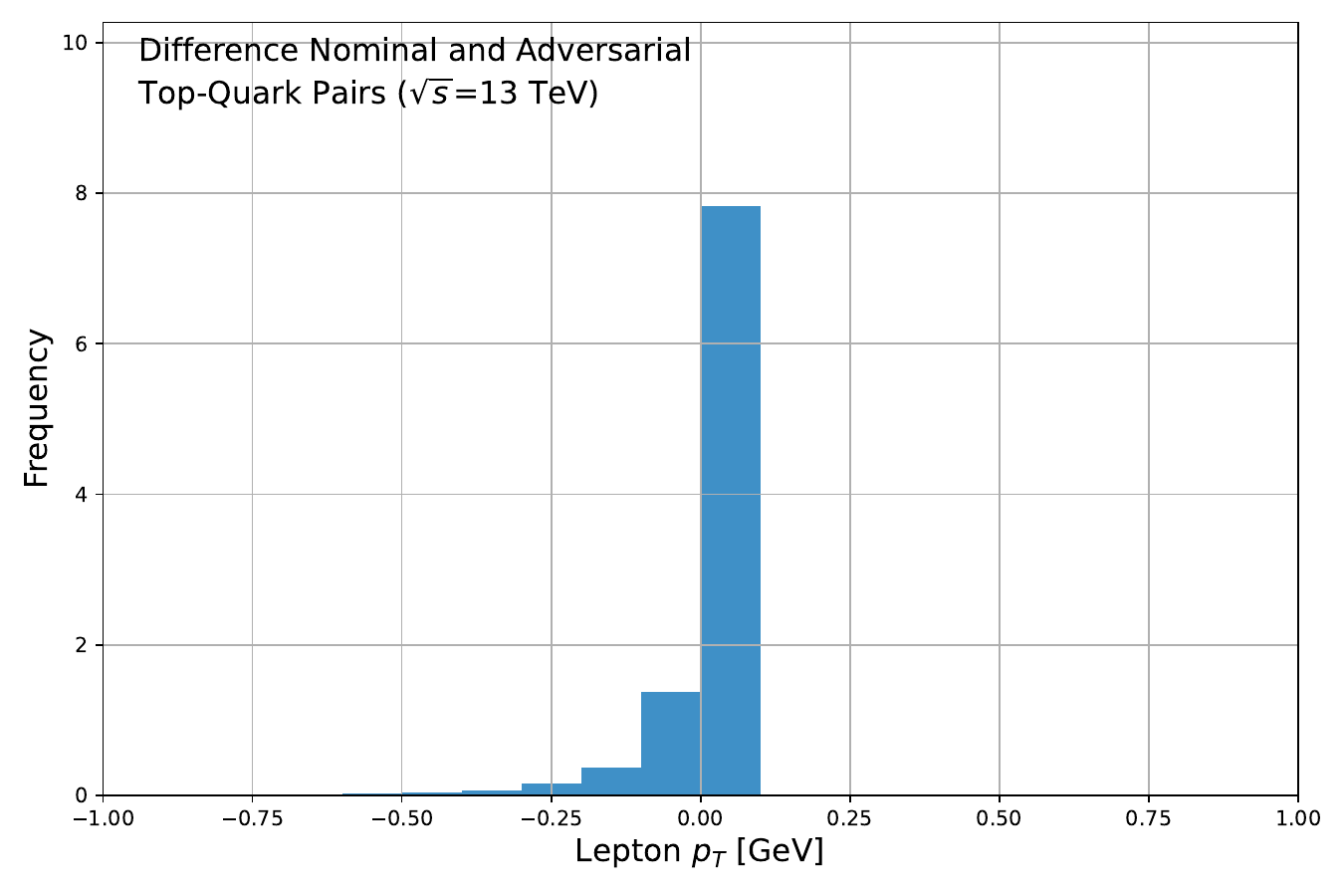}
    \includegraphics[width=0.32\textwidth]{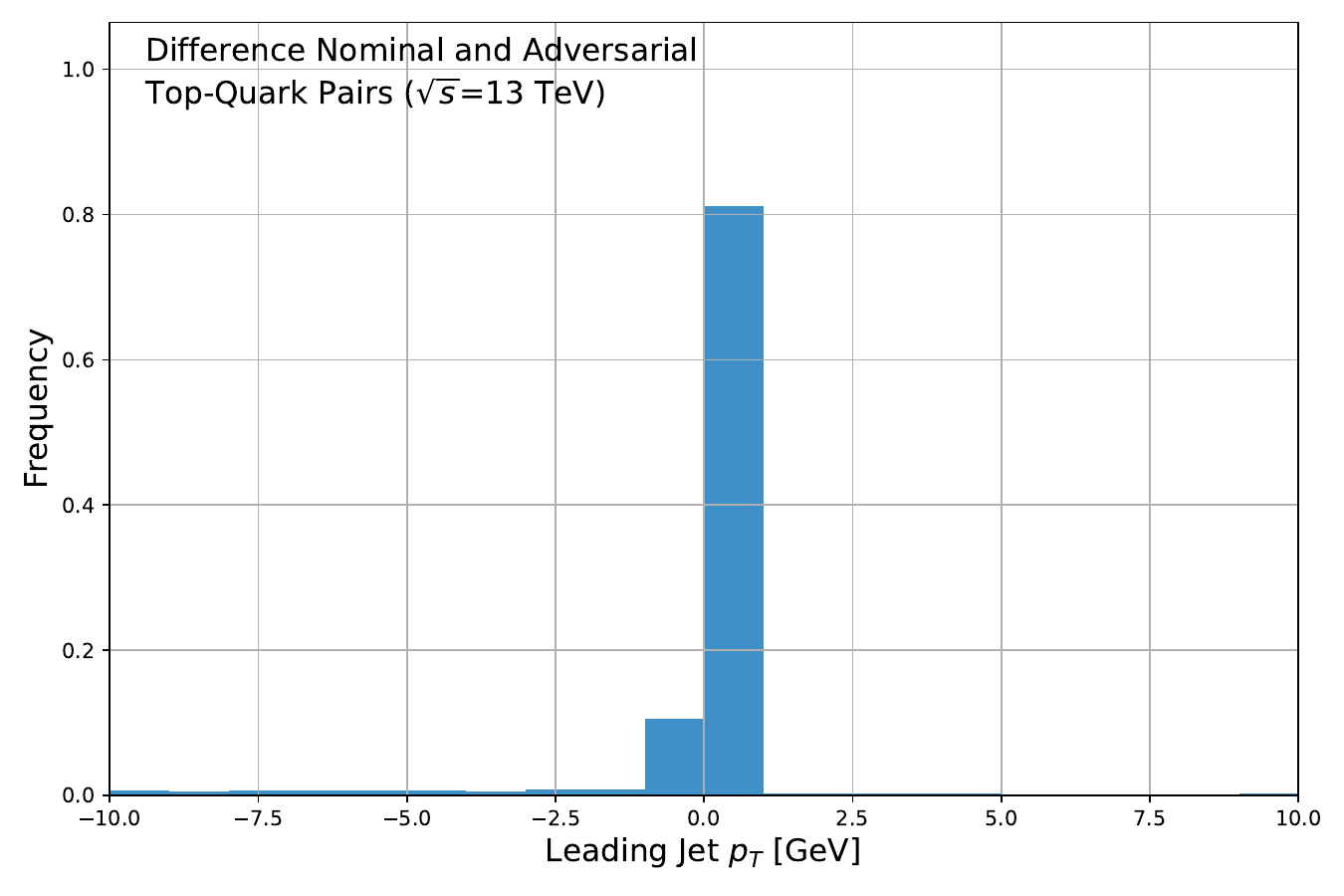}
    \includegraphics[width=0.32\textwidth]{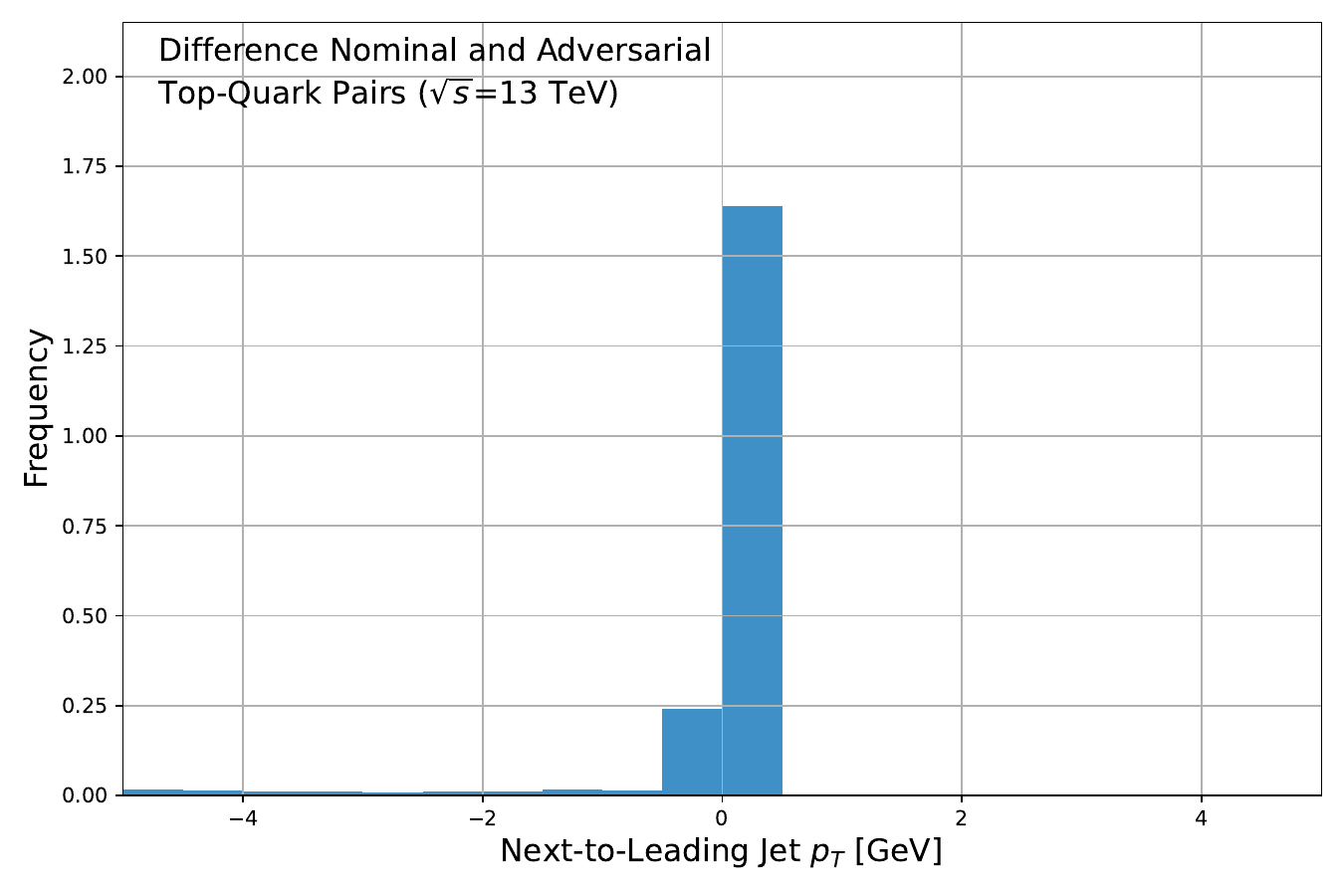}
\caption{Results for the $t\bar{t}$ versus $WW$ classifier using the alternative (C\&W) loss function Eqn. \ref{eqn:CW}. 
Top row: Comparison of the nominal (yellow) and adversarial (blue) distributions for three representative input observables in $t\bar{t}$ events: the transverse momentum of the reconstructed lepton (left), the transverse momentum of the leading jet (middle), and the transverse momentum of the subleading jet (right). Bottom row: event-by-event differences between the nominal and adversarial distributions for the same observables. \label{fig:adverWWTopCW}}
\end{figure}

When repeating the study using the C\&W-inspired loss function, we obtain results that are qualitatively very similar to those observed with the nominal cross-entropy-based approach. In particular, we find a slightly larger degradation of the signal efficiency on the adversarial dataset, while the impact on the overall ROC AUC remains somewhat smaller; both effects are consistent within statistical uncertainties. The induced feature shifts and perturbation patterns, shown in Fig.~\ref{fig:adverWWTopCW}, closely resemble those obtained with the nominal method, indicating that both constructions probe similar directions in feature space.
Varying the adversarial hyperparameters, in particular the margin parameter $\kappa$, allows the fooling ratio to be tuned. However, increasing $\kappa$ typically leads to more pronounced deformations of individual features, which can result in deviations from the imposed physical constraints and reduced agreement with the nominal distributions. This reflects the inherent trade-off between adversarial strength and physical realism in the hybrid formulation.
Given the close agreement between the results obtained with the nominal loss function in Eq.~(\ref{eqn:crossattack}) and the alternative C\&W-inspired loss in Eq.~(\ref{eqn:CW}), we do not consider the latter further in the subsequent examples.

\subsection{Graph Neural Network Based Classifiers}

As a representative example of a high-capacity, low-level model, we consider the graph neural network (GNN)–based quark–gluon jet tagger described above. The architecture consists of three stacked EdgeConv blocks followed by a fully connected classification head, allowing the model to learn hierarchical local and global correlations among jet constituents. In total, the network comprises approximately 100k trainable parameters, significantly exceeding the capacity of the MLP benchmark. Architectures of comparable complexity are widely used in modern jet-substructure studies and constitute a realistic example of state-of-the-art particle-level learning at the LHC.

The quark--gluon tagging study is performed on a dataset of 50\,000 simulated jets. Adversarial examples are again generated using the uncertainty-constrained projected gradient descent algorithm described in Chapter~\ref{sec:framework_attack}, applied at the track level. The optimization is performed with a step size $\alpha = 0.04$ and 20 iterations, while the maximal per-feature deviation is restricted to $3\sigma$. The regularization strengths are set to $\lambda_{\Delta}=0.5$ and $\lambda_{\chi^2}=0.5$. A feature mask is applied such that only physically meaningful observables are perturbed.

Per-feature uncertainties are defined relative to the observable magnitude via $\sigma_i = f_i |x_i|$, with fractional widths of 2\% for the transverse momentum, 0.1\% for the angular variables $\Delta\eta$ and $\Delta\phi$, 0.5\% for the impact parameters $d_0$ and $z_0$, and no variation applied to the electric charge. These values reflect realistic detector resolutions at track level and ensure that perturbations remain within experimentally allowed ranges.

\begin{figure}[htbp]
    \centering
    \includegraphics[width=0.32\textwidth]{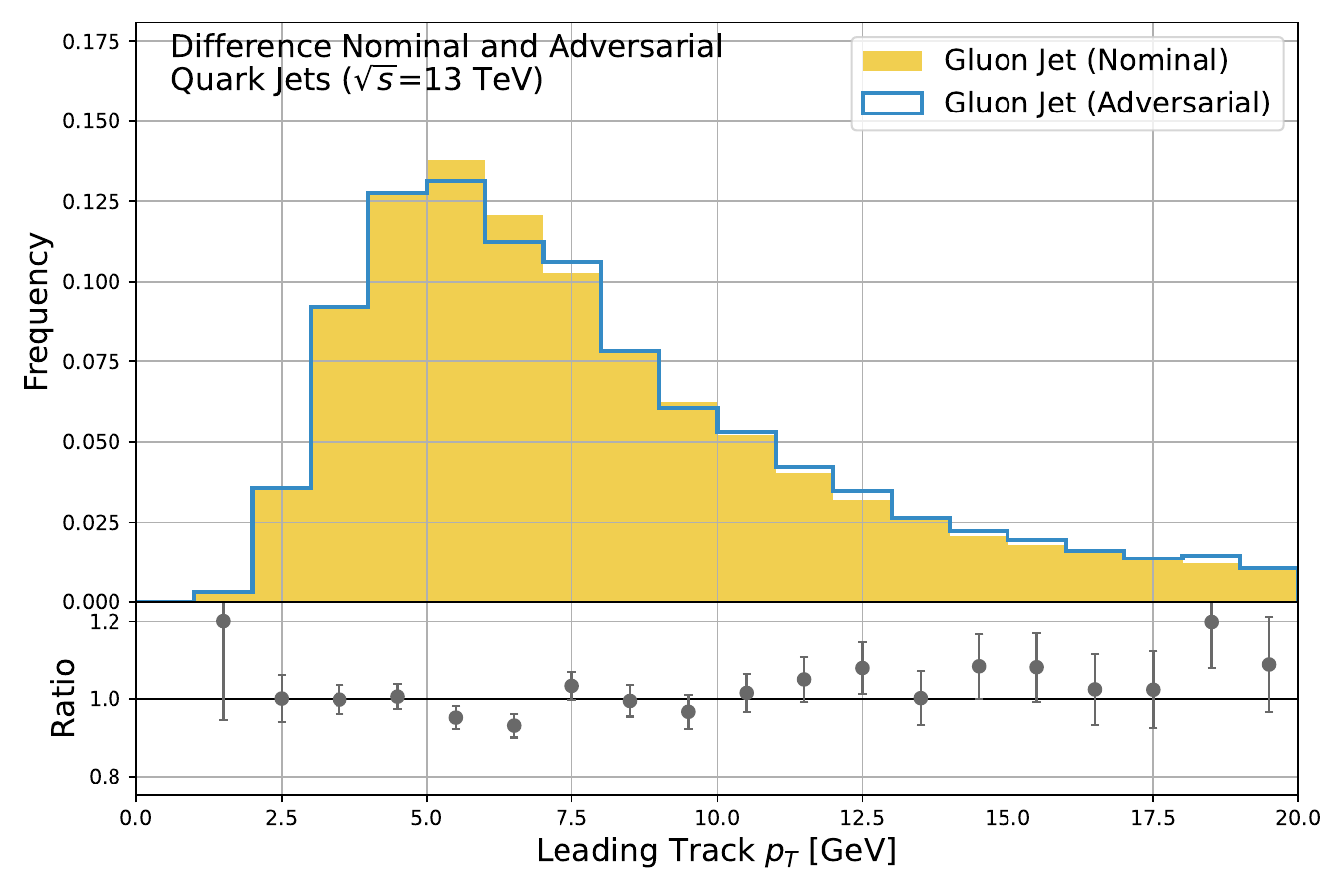}
    \includegraphics[width=0.32\textwidth]{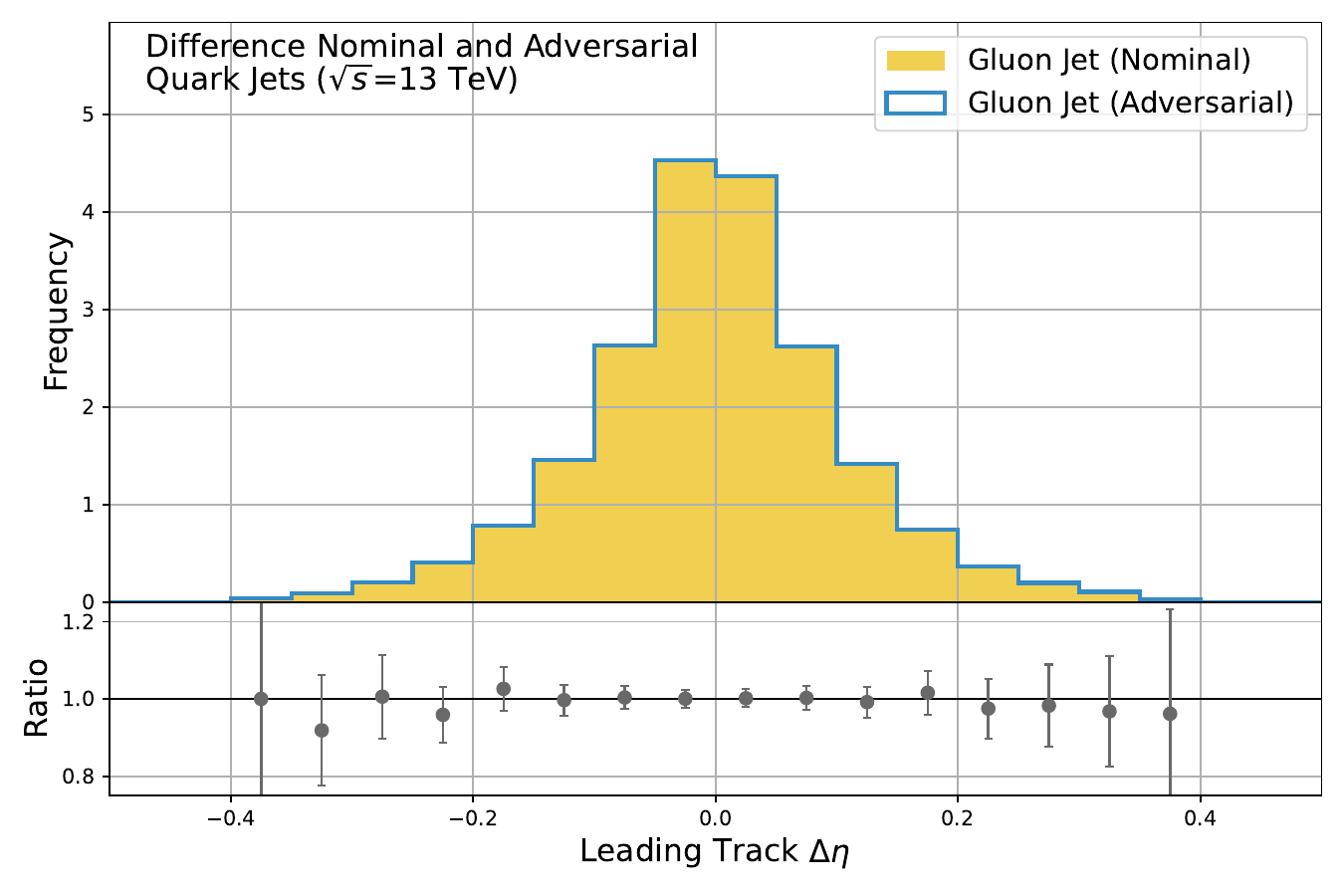}
    \includegraphics[width=0.32\textwidth]{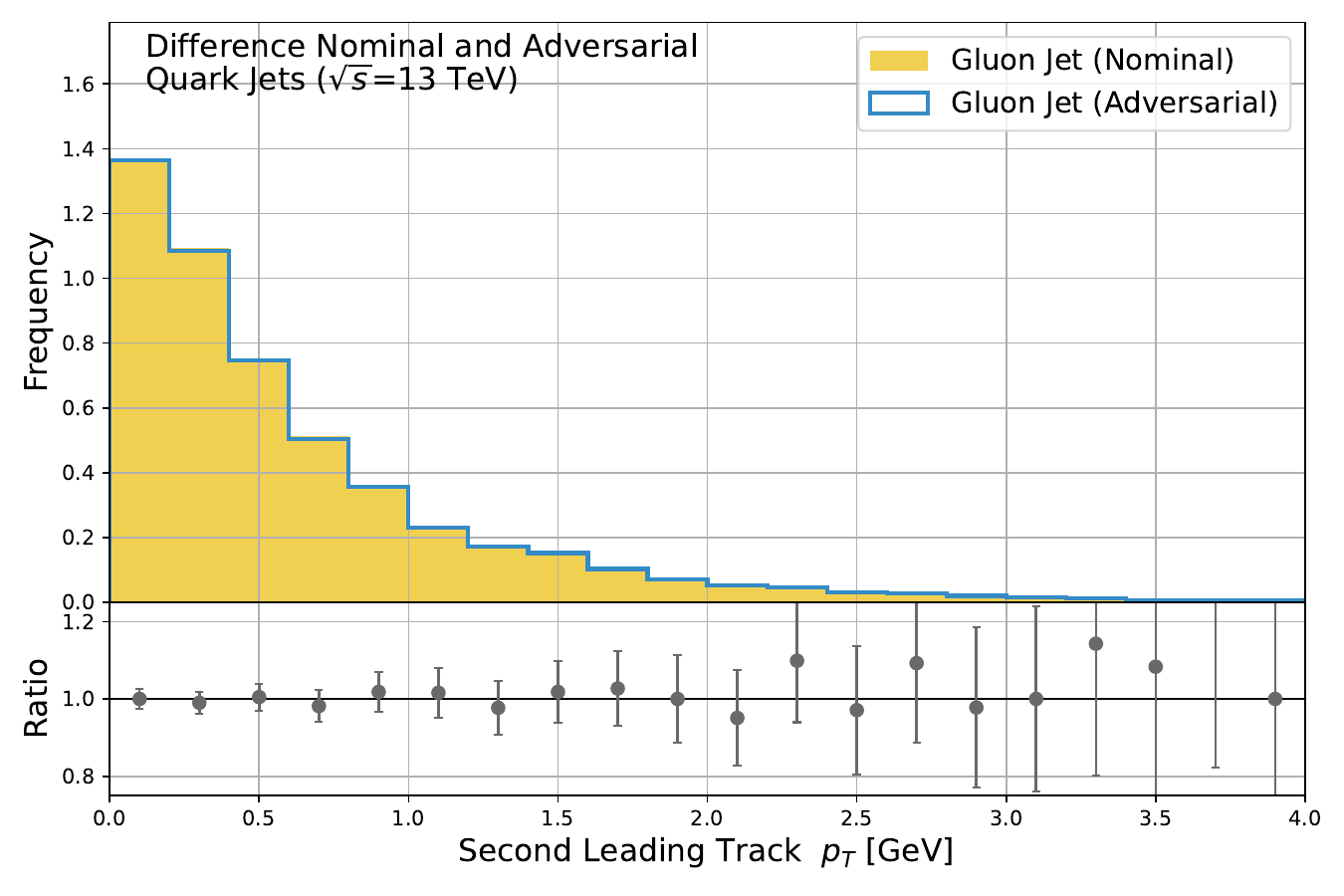}

    \includegraphics[width=0.32\textwidth]{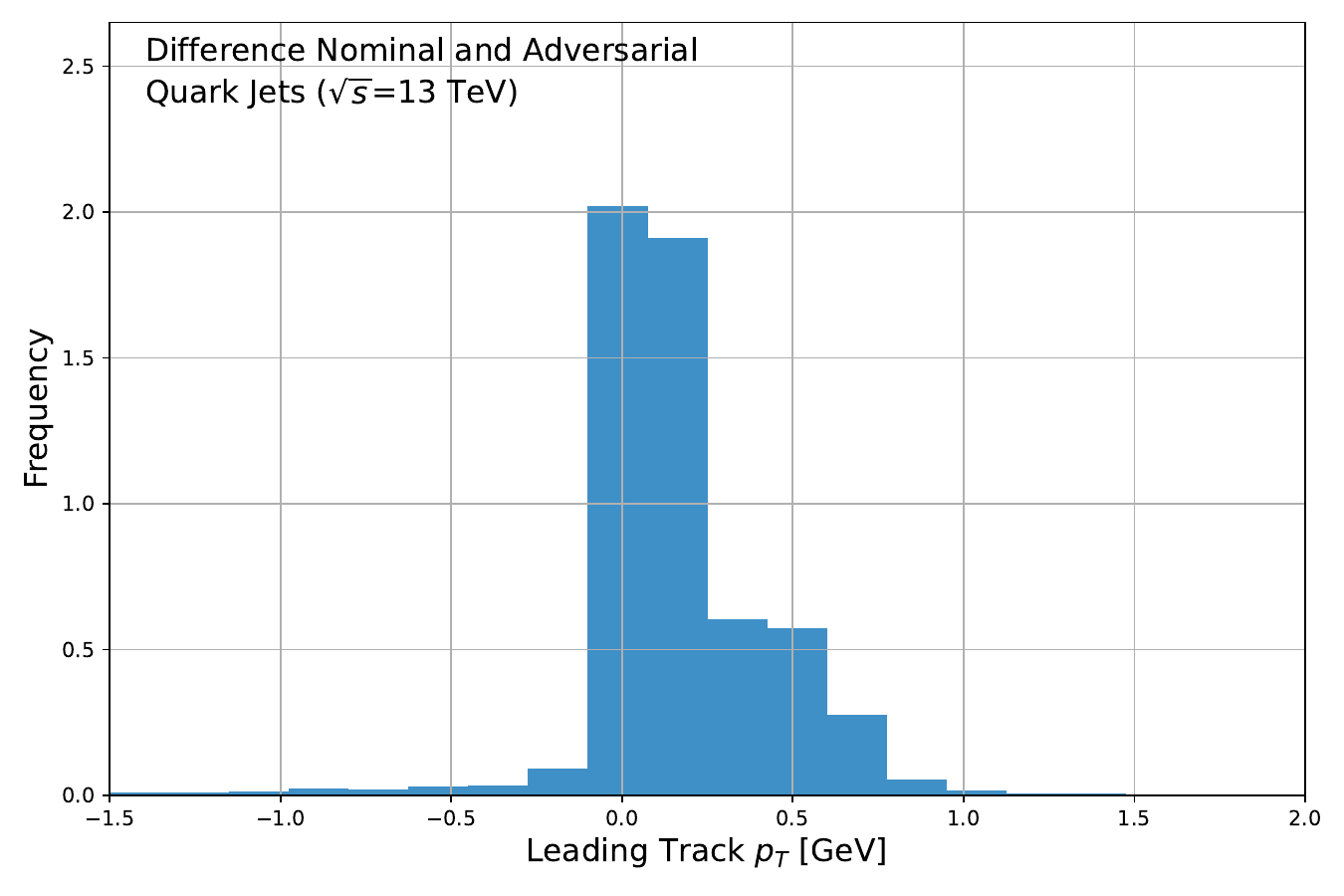}
    \includegraphics[width=0.32\textwidth]{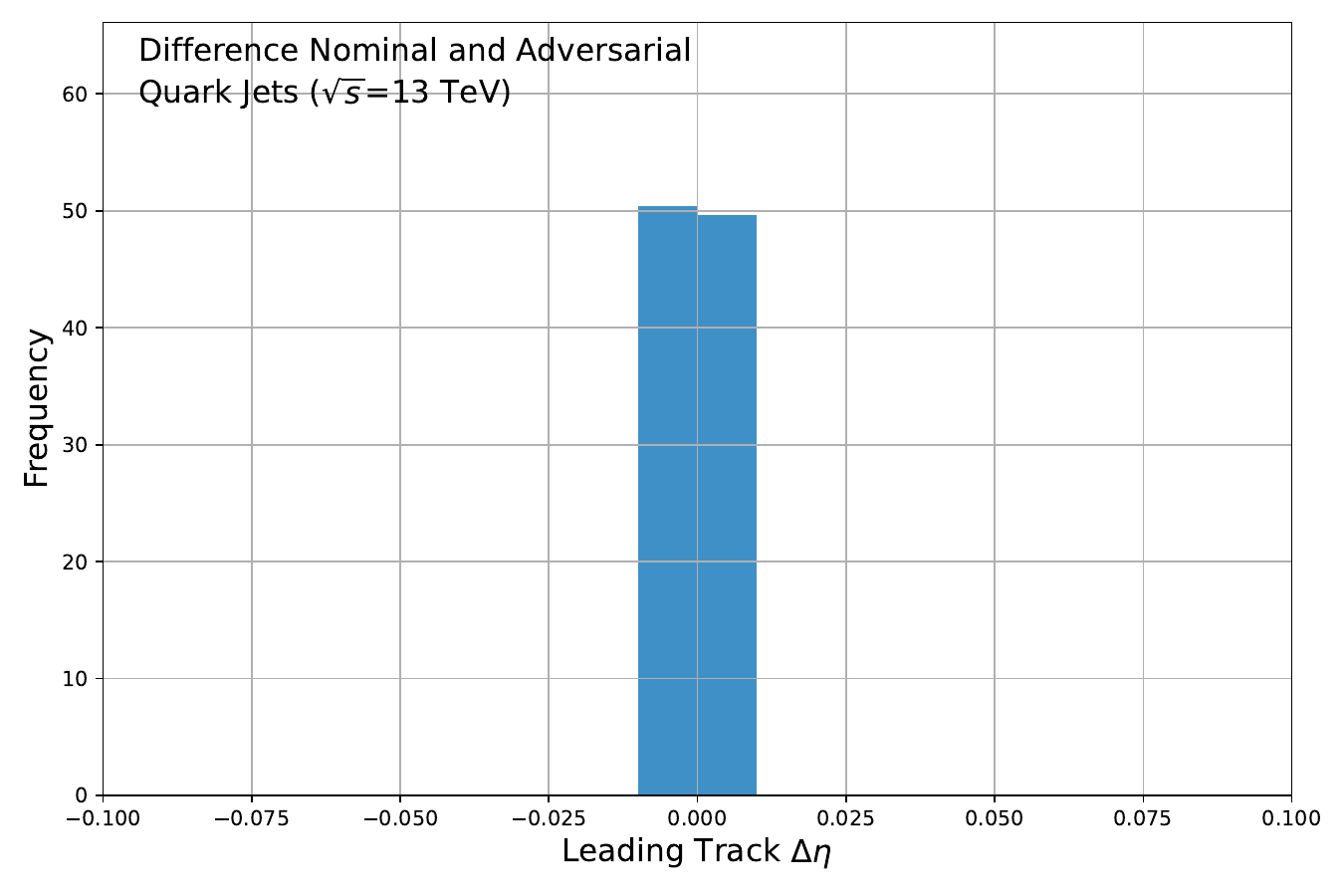}
    \includegraphics[width=0.32\textwidth]{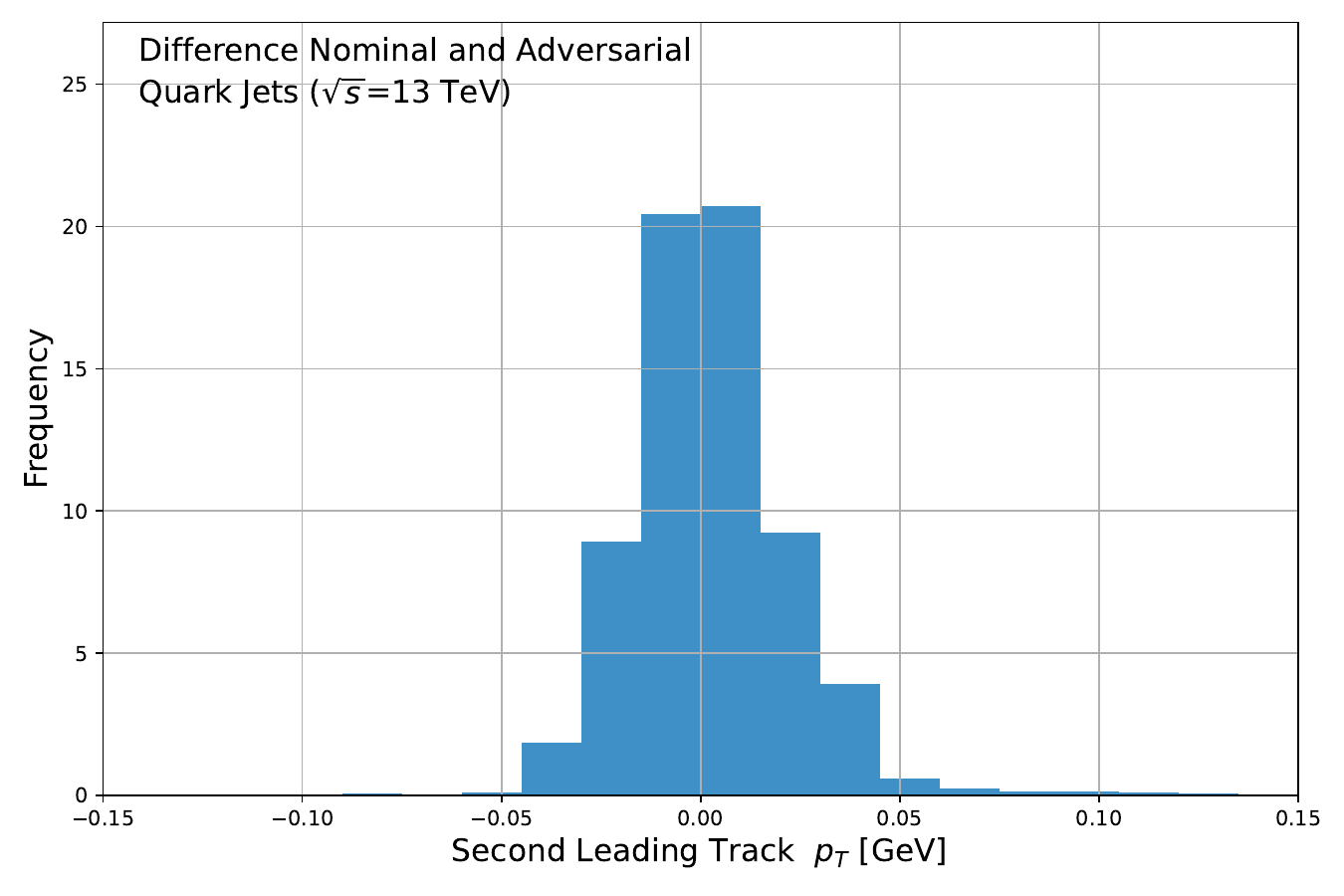}
\caption{Results for the quark--gluon jet tagging classifier. 
Top row: Comparison of the nominal (yellow) and adversarial (blue) distributions for three representative track-level observables in signal jets: the transverse momentum of the leading track (left), the pseudorapidity difference between the leading track and the jet axis (middle), and the transverse momentum of the second-leading track (right). 
Bottom row: event-by-event differences between the nominal and adversarial distributions for the same observables, demonstrating the minimal distortions induced by the adversarial perturbations. 
\label{fig:adverJetTagger}}
\end{figure}

The adversarial results for the GNN-based jet tagger are shown in Figure~\ref{fig:adverJetTagger}. The top row compares the nominal (yellow) and adversarial (blue) distributions for three representative track-level observables, while the bottom row displays the corresponding bin-by-bin differences. For the leading-track transverse momentum, the difference exhibits the expected approximately Gaussian structure induced by the prior constraint. In contrast, no visible modification is observed for $\Delta\eta$, reflecting the very small allowed angular uncertainty. For the second-leading track transverse momentum, a Gaussian-like shift is again visible but smaller in magnitude, consistent with the relative definition of the uncertainty and the typically lower transverse momentum of subleading tracks. Overall, the distortions remain within statistical uncertainties and are not identifiable by standard distributional checks. Furthermore, the Pearson correlation matrices of the input features show no significant deviation between nominal and adversarial samples.

The results for the GNN-based quark--gluon tagger are summarized in Table~\ref{tab:adversarial_comparison_fullset}. Similar to the previous test-case, the cut-based baseline shows only a very small sensitivity to the adversarial perturbations, with changes in signal efficiency and background rejection below the percent level and consistent with statistical fluctuations. 

In contrast, the GNN trained on nominal data exhibits a clear performance degradation when evaluated on adversarial samples. The signal efficiency at fixed background rejection decreases by $0.013 \pm 0.003$, and the ROC AUC is reduced by $0.012 \pm 0.003$, indicating sensitivity to high-dimensional perturbations that remain invisible in individual feature distributions.

\begin{table}[htbp]
\centering
\caption{Performance comparison for the quark/gluon jet tagger on the full dataset between nominal and adversarial evaluation samples for different training strategies.}
\label{tab:adversarial_comparison_fullset}
\begin{tabular}{l|ccc}
\hline
& Nominal Eval. & Adversarial Eval. & Difference \\
\hline
\multicolumn{4}{l}{\textbf{Baseline Performance (Cut-based)}} \\
\hline
Signal Efficiency & $0.571 \pm 0.004$ & $0.563 \pm 0.004$ & $0.008 \pm 0.005$ \\
Background Rejection & $0.491 \pm 0.001$ & $0.484 \pm 0.004$ & $0.007 \pm 0.004$ \\
\hline
\multicolumn{4}{l}{\textbf{DNN -- Trained on Nominal Sample Only}} \\
\hline
Signal Efficiency (at fixed background rejection) & $0.794 \pm 0.002$ & $0.781 \pm 0.002$ & $0.013 \pm 0.003$ \\
ROC AUC & $0.704 \pm 0.002$ & $0.692 \pm 0.002$ & $0.012 \pm 0.003$ \\
\hline
\multicolumn{4}{l}{\textbf{DNN -- Trained on Adversarial Sample Only}} \\
\hline
Signal Efficiency (at fixed background rejection) & $0.779 \pm 0.003$ & $0.781 \pm 0.002$ & $-0.003 \pm 0.004$ \\
ROC AUC & $0.688 \pm 0.005$ & $0.698 \pm 0.007$ & $-0.011 \pm 0.008$ \\
\hline
\multicolumn{4}{l}{\textbf{DNN -- Trained on Nominal + Adversarial Sample}} \\
\hline
Signal Efficiency (at fixed background rejection) & $0.806 \pm 0.003$ & $0.802 \pm 0.002$ & $0.004 \pm 0.004$ \\
ROC AUC & $0.716 \pm 0.004$ & $0.714 \pm 0.003$ & $0.002 \pm 0.005$ \\
\hline
\end{tabular}
\end{table}

Training exclusively on adversarial samples reduces the performance difference between nominal and adversarial evaluation, while combined training on nominal and adversarial data largely stabilizes the classifier. In this case, the residual differences are statistically compatible with zero. Overall, the results confirm a non-negligible sensitivity of the graph-based model to physically allowed perturbations, which can be mitigated through adversarial exposure during training.

\subsection{Transformer Based Classifiers}

As a representative example of a high-capacity, set-based architecture, we consider the transformer-based $E_{\mathrm{T}}^{\mathrm{miss}}$ classifier introduced above. The model processes variable-length sequences of charged-particle tracks using a learnable embedding layer followed by three stacked transformer encoder blocks with multi-head self-attention. A classification head maps the aggregated representation to a binary output. In total, the network comprises approximately 400ke parameters and is trained on a dataset of 150\,000 simulated events. This architecture represents the most expressive model studied in this work and allows us to probe adversarial sensitivity in a regime where complex global correlations among all tracks can be exploited.

Adversarial examples for the transformer-based $E_{\mathrm{T}}^{\mathrm{miss}}$ classifier are generated with a step size $\alpha = 0.02$ over 15 iterations. The per-feature uncertainties are defined relative to the observable magnitude via $\sigma_i = f_i |x_i|$, with fractional widths $f_i = (0.04,\, 0.001,\, 0.04,\, 0.002)$ corresponding to the track-level features $(p_x, p_y, p_z, d_0)$. The transverse momentum components $p_x$ and $p_z$ are allowed to vary at the 4\% level, while the impact parameter $d_0$ is varied at the 0.2\% level. The variation of $p_y$ is intentionally restricted to a negligible level (0.1\%) to avoid introducing artificial distortions in the reconstructed transverse momentum balance, since $p_y$ is strongly constrained by the measured transverse momentum and correlated with $p_x$. 

\begin{figure}[htbp]
    \centering
    \includegraphics[width=0.32\textwidth]{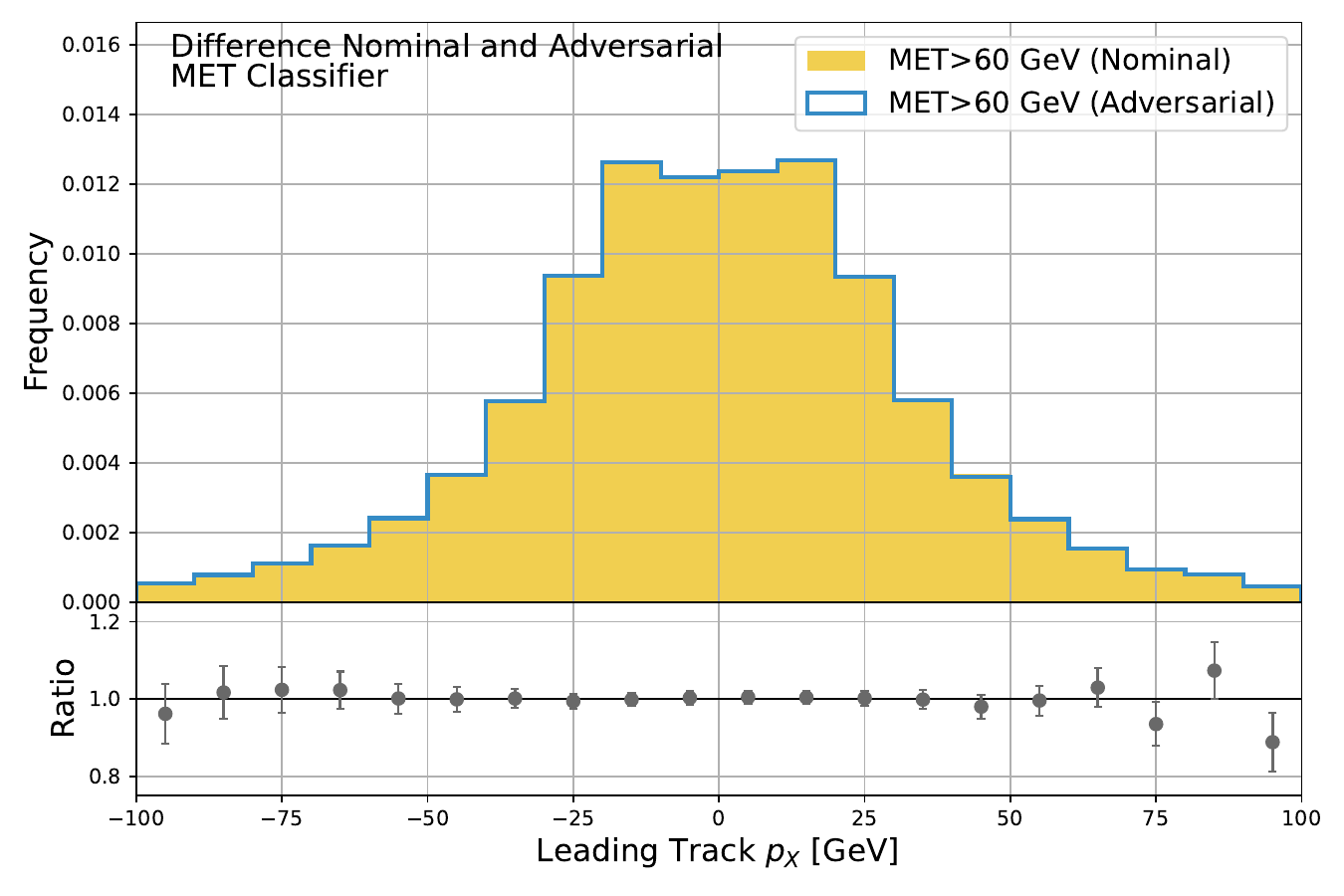}
    \includegraphics[width=0.32\textwidth]{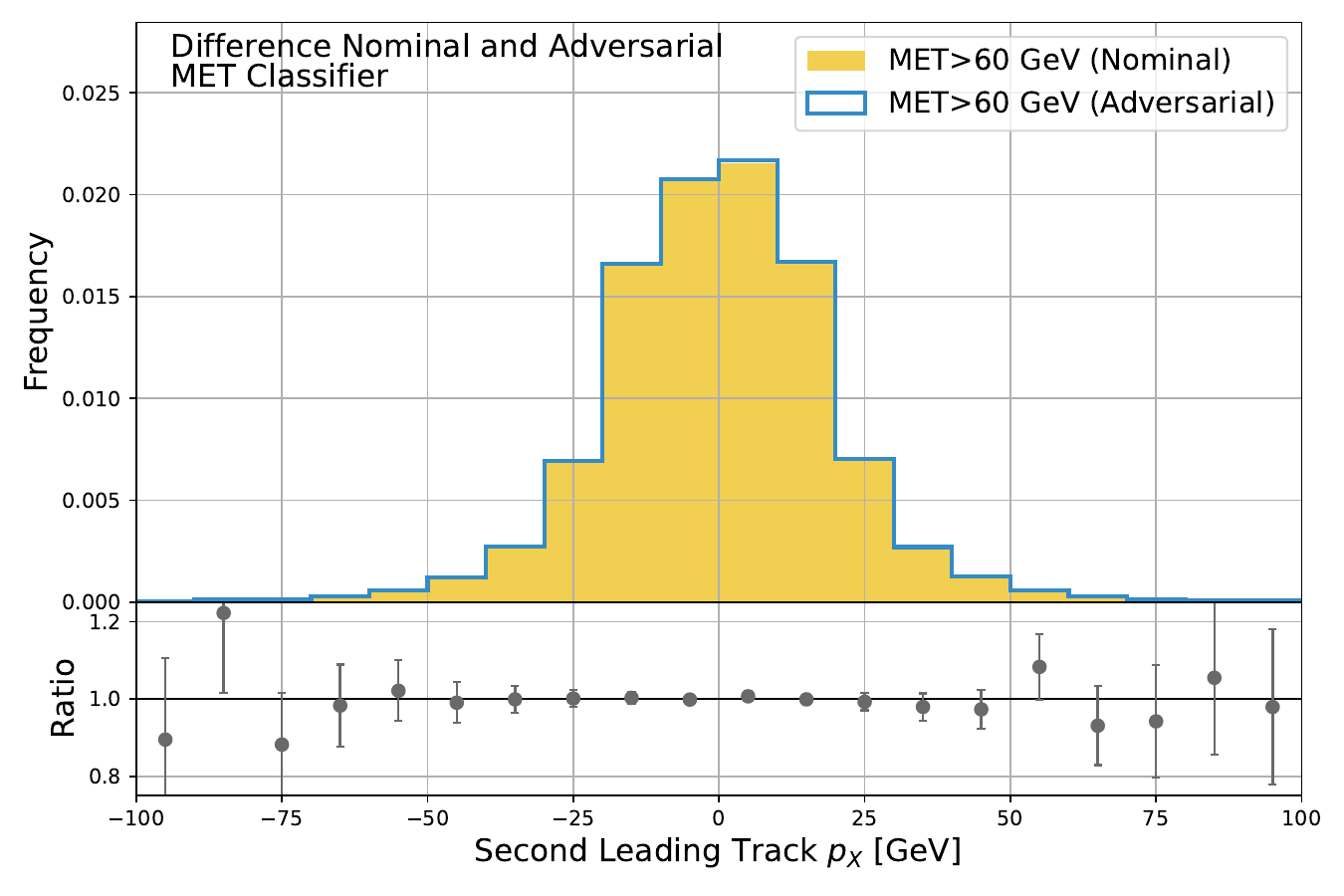}
    \includegraphics[width=0.32\textwidth]{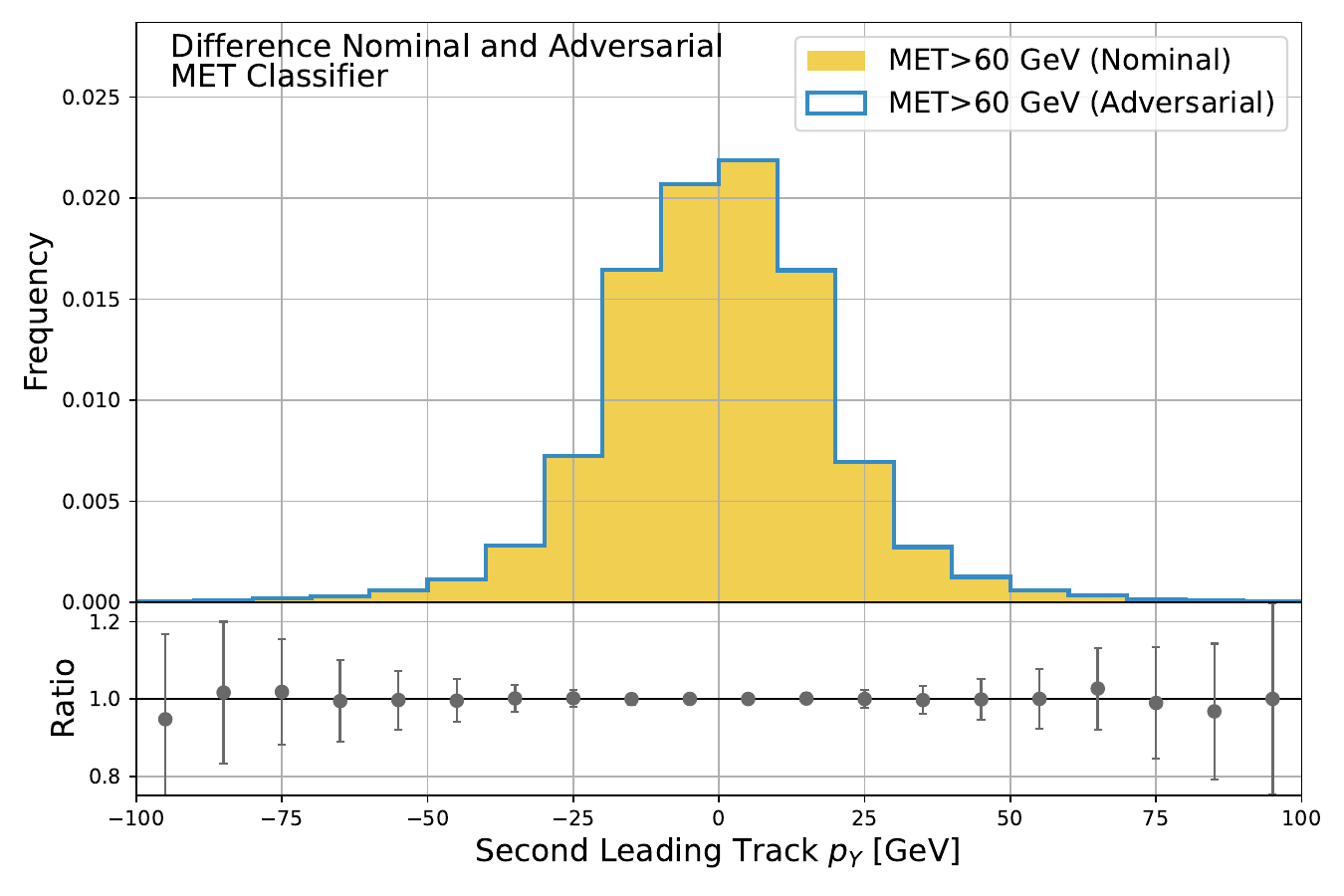}

    \includegraphics[width=0.32\textwidth]{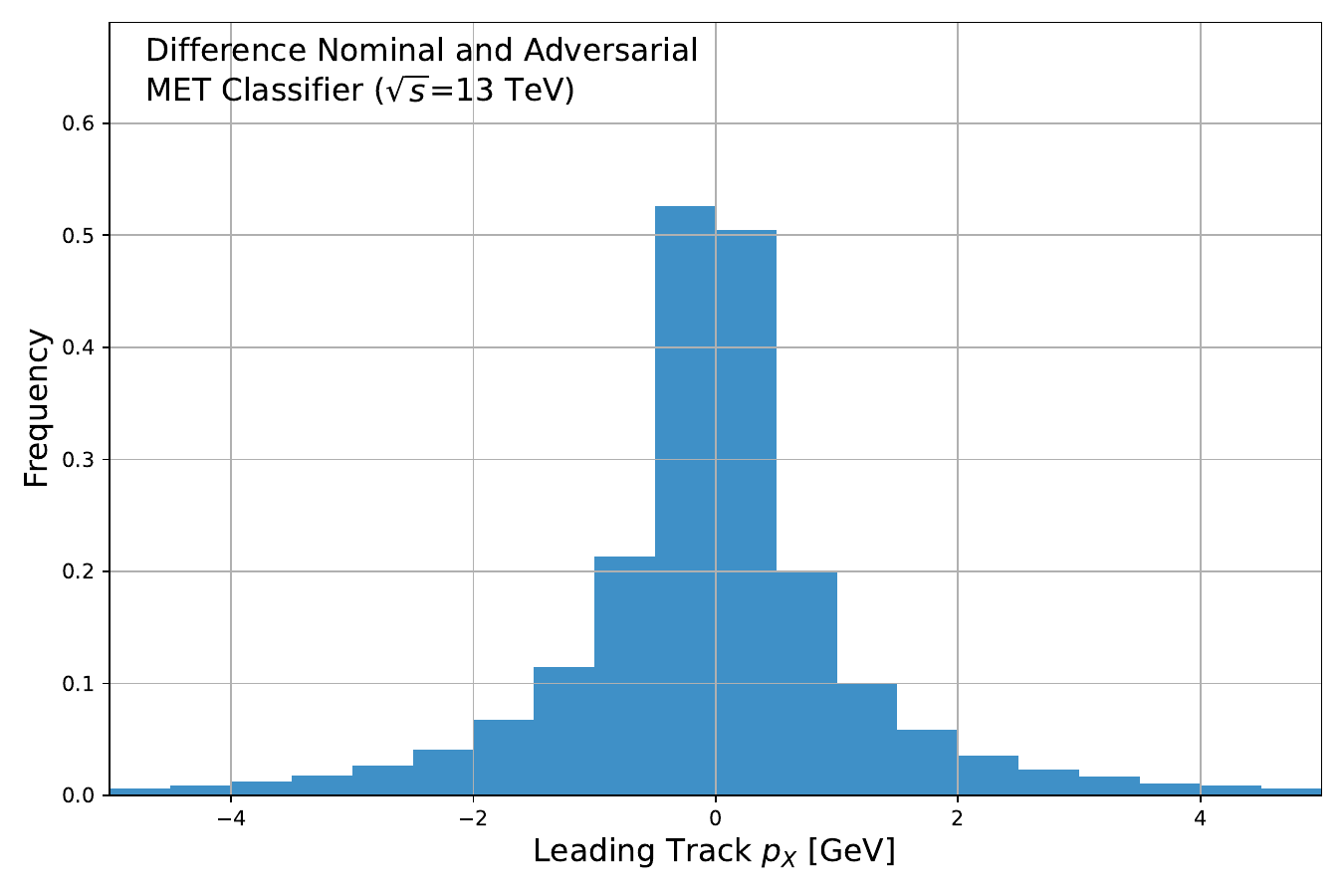}
    \includegraphics[width=0.32\textwidth]{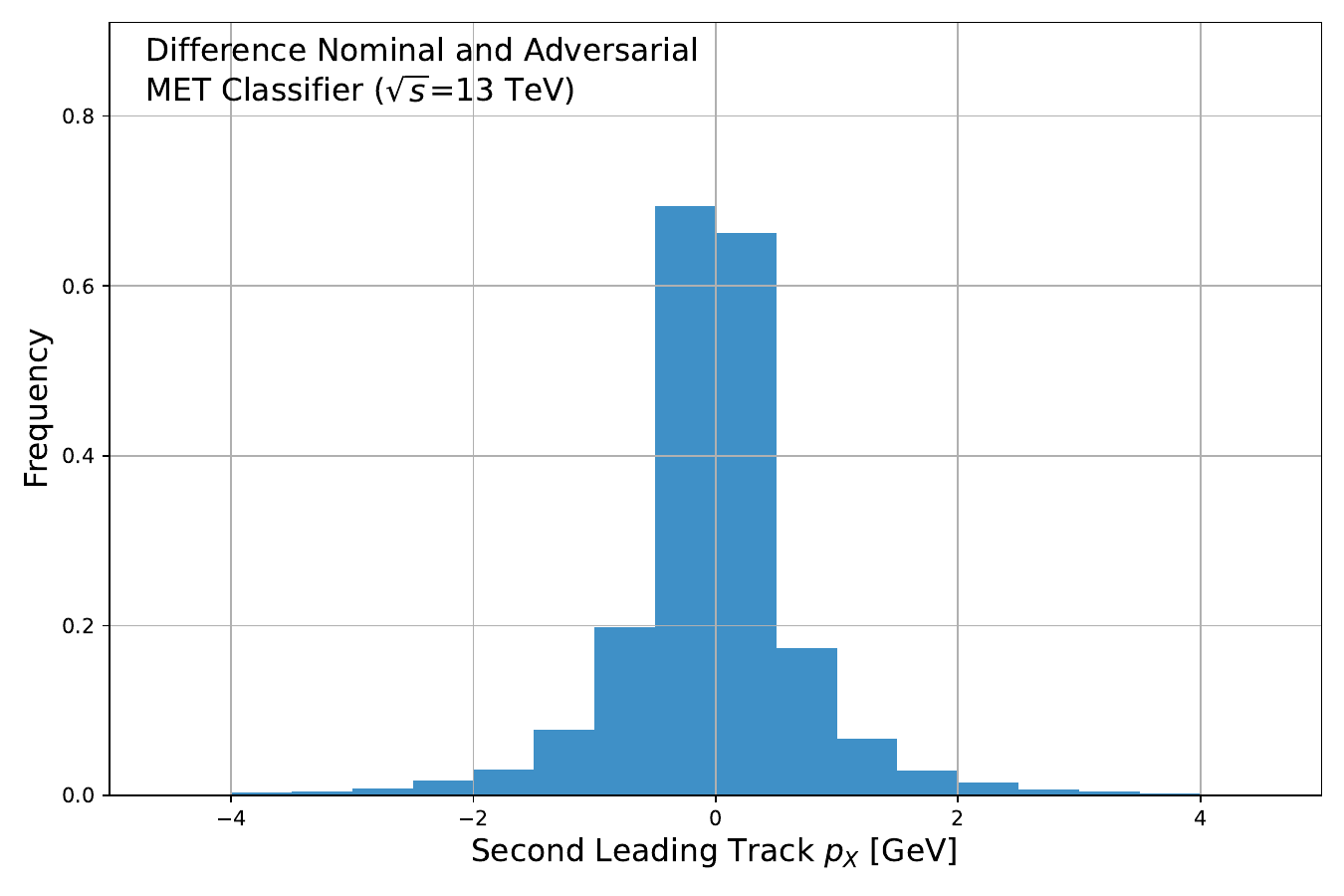}
    \includegraphics[width=0.32\textwidth]{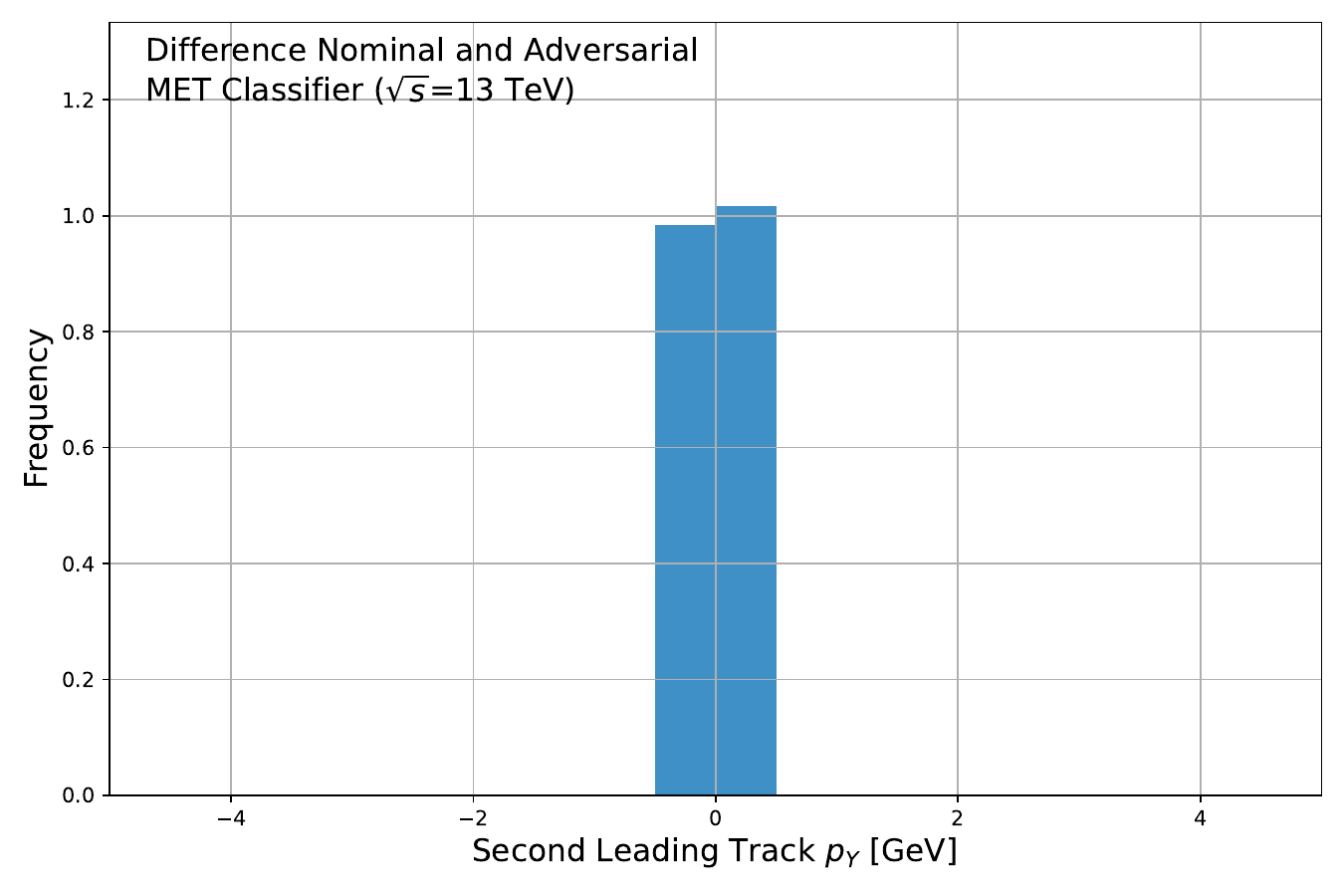}
    \caption{Results for the transformer-based $E_{\mathrm{T}}^{\mathrm{miss}}$ classifier.  Top row: Comparison of the nominal (yellow) and adversarial (blue) distributions for three representative track-level observables in signal events: the $p_x$ component of the leading track (left), and the $p_x$ (middle) and $p_y$ (right) components of the second-leading track.  Bottom row: event-by-event differences between the nominal and adversarial distributions for the same observables. The perturbations follow the imposed Gaussian prior and remain within statistical uncertainties, demonstrating that no significant distortions of the one-dimensional feature distributions are introduced. 
\label{fig:adverETMiss}}
\end{figure}

Figure~\ref{fig:adverETMiss} illustrates the comparison between nominal and adversarial feature distributions for representative track-level observables in the $E_{\mathrm{T}}^{\mathrm{miss}}$ task. The top row shows the nominal and adversarial distributions, while the bottom row displays their bin-by-bin differences. As expected from the imposed Gaussian prior and per-feature uncertainty constraints, only small, symmetric distortions are observed. In all cases, the deviations remain within statistical uncertainties, indicating that the adversarial perturbations do not introduce visible changes in the one-dimensional feature distributions. Also no significant difference in the Pearson correlation coefficients among the input features is observed. 

The results for the transformer-based $E_{\mathrm{T}}^{\mathrm{miss}}$ classifier are summarized in Table~\ref{tab:adversarial_comparison}. The cut-based baseline shows only a negligible sensitivity to the adversarial perturbations, with changes in signal efficiency and background rejection at the level of a few per mille and consistent with statistical fluctuations.

For the transformer model trained exclusively on nominal data, a clear degradation is observed when evaluated on adversarial samples. The signal efficiency at fixed background rejection decreases by $0.011 \pm 0.004$, and the ROC AUC is reduced by $0.010 \pm 0.004$. This indicates that also the high-capacity architecture is sensitive to small but correlated perturbations that remain invisible in individual feature distributions as well as one-dimensional correlations.

Training solely on adversarial samples does not eliminate this effect; performance differences between nominal and adversarial evaluation remain at the percent level. In contrast, training on the combined nominal and adversarial dataset largely stabilizes the classifier, with the remaining differences being statistically compatible with zero. 

Overall, the results confirm that the transformer-based model exhibits a measurable sensitivity to physically allowed perturbations, while adversarial exposure during training substantially mitigates the observed performance shift.

\begin{table}[htbp]
\centering
\caption{Performance comparison for the $E^T_{Miss}$ classifier between nominal and adversarial evaluation samples for different training strategies.}
\label{tab:adversarial_comparison}
\begin{tabular}{l|ccc}
\hline
 & Nominal Eval. & Adversarial Eval. & Difference \\
\hline
\multicolumn{4}{l}{\textbf{Baseline Performance (Cut-based)}} \\
\hline
Signal Efficiency & $0.405 \pm 0.002$ & $0.401 \pm 0.002$ & $0.004 \pm 0.003$ \\
Background Rejection & $0.747 \pm 0.002$ & $0.744 \pm 0.002$ & $0.003 \pm 0.003$ \\
\hline
\multicolumn{4}{l}{\textbf{DNN -- Trained on Nominal Sample Only}} \\
\hline
Signal Efficiency (at fixed background rejection) & $0.490 \pm 0.003$ & $0.480 \pm 0.003$ & $0.011 \pm 0.004$ \\
ROC AUC & $0.668 \pm 0.003$ & $0.658 \pm 0.003$ & $0.010 \pm 0.004$ \\
\hline
\multicolumn{4}{l}{\textbf{DNN -- Trained on Adversarial Sample Only}} \\
\hline
Signal Efficiency (at fixed background rejection) & $0.479 \pm 0.004$ & $0.464 \pm 0.003$ & $0.014 \pm 0.005$ \\
ROC AUC & $0.657 \pm 0.004$ & $0.648 \pm 0.003$ & $0.010 \pm 0.005$ \\
\hline
\multicolumn{4}{l}{\textbf{DNN -- Trained on Nominal + Adversarial Sample}} \\
\hline
Signal Efficiency (at fixed background rejection) & $0.480 \pm 0.003$ & $0.488 \pm 0.003$ & $-0.008 \pm 0.004$ \\
ROC AUC & $0.660 \pm 0.003$ & $0.666 \pm 0.003$ & $-0.006 \pm 0.004$ \\
\hline
\end{tabular}
\end{table}

\subsection{Interpretation of the Results}

The results presented in the previous sections consistently demonstrate that all DNN-based classifiers exhibit a measurable performance degradation at the percent level when evaluated on adversarial samples, while the corresponding cut-based baselines remain largely unaffected. Although the distortions introduced by the uncertainty-constrained perturbations are statistically indistinguishable in one-dimensional feature distributions and correlation matrices, they nevertheless induce non-negligible shifts in the decision boundaries of high-capacity models. Interestingly, the relative degradation is largest for the smallest network architecture, whereas the more expressive GNN and transformer models show somewhat reduced sensitivity. This trend suggests that vulnerability is not simply driven by parameter count, but rather by how strongly a model relies on a limited set of dominant features versus a more distributed and redundant internal representation.

What changes the performance is not a visible distortion of the underlying feature distributions, but rather small, targeted shifts of individual events in feature space. In particular, events that lie close to the decision boundary of the network are mildly perturbed such that they cross into the opposite class. This interpretation is supported by dedicated studies of the misclassified events, which show that they are predominantly located near the classifier’s decision threshold in the output space. Consequently, even perturbations that are statistically indistinguishable at the distribution level can induce a measurable degradation in performance by systematically reassigning these boundary events.

The observed behavior is consistent with expectations. Deep neural networks exploit subtle, high-dimensional correlations among input features and learn comparatively complex decision boundaries that are finely adapted to structures present in the training data. Small but coherent perturbations within experimentally allowed uncertainties can therefore shift events across these boundaries, even when no significant distortions are visible in one-dimensional feature distributions. In contrast, cut-based selections rely on a limited number of independent threshold requirements and define geometrically simple decision regions that are inherently less sensitive to correlated variations. The percent-level degradation observed for all DNN-based classifiers, significantly larger than for the cut-based baselines, is thus consistent with the increased expressive power, and corresponding sensitivity, of high-capacity models.

While one might naively expect larger and more expressive networks to be more vulnerable to adversarial perturbations, the observed pattern can be understood from the structure of the learned representation rather than from parameter count alone. The smallest model in our study, the fully connected MLP used for the $t\bar{t}$ versus $WW$ task, operates on a limited set of high-level observables. Its decision boundary therefore relies strongly on a relatively small number of dominant kinematic features. When coherent, uncertainty-constrained perturbations are applied directly to these observables, the classifier has little internal redundancy to compensate for such shifts. As a result, even small but correlated deformations in feature space can lead to a comparatively larger degradation in performance. In this sense, low-dimensional models depending on a compact set of engineered variables are naturally more sensitive to structured variations along physically relevant directions.

In contrast, the GNN and transformer architectures learn distributed representations over many low-level inputs. Their predictions emerge from aggregating information across numerous tracks and from modeling complex local and global correlations. Small Gaussian perturbations applied to individual inputs are therefore effectively diluted across a larger representational space. This introduces a form of redundancy: the decision boundary is supported by many correlated features rather than a few dominant ones. Consequently, the relative impact of the same uncertainty-constrained perturbation can be reduced, even though the models are more expressive and contain significantly more parameters. The results therefore suggest that adversarial sensitivity is governed less by model size and more by how concentrated or distributed the learned representation is within the feature space.

\section{Estimation of Hidden Systematic Uncertainties}

The observed percent-level performance shifts under physically allowed and statistically indistinguishable perturbations indicate the presence of a model-dependent uncertainty intrinsic to neural-network based classifiers. While this effect can be mitigated through adversarial exposure during training, it is not captured by conventional validation procedures and therefore constitutes an additional source of systematic sensitivity in DNN-based analyses.

\begin{figure}[htbp]
    \centering
    \includegraphics[width=0.79\textwidth]{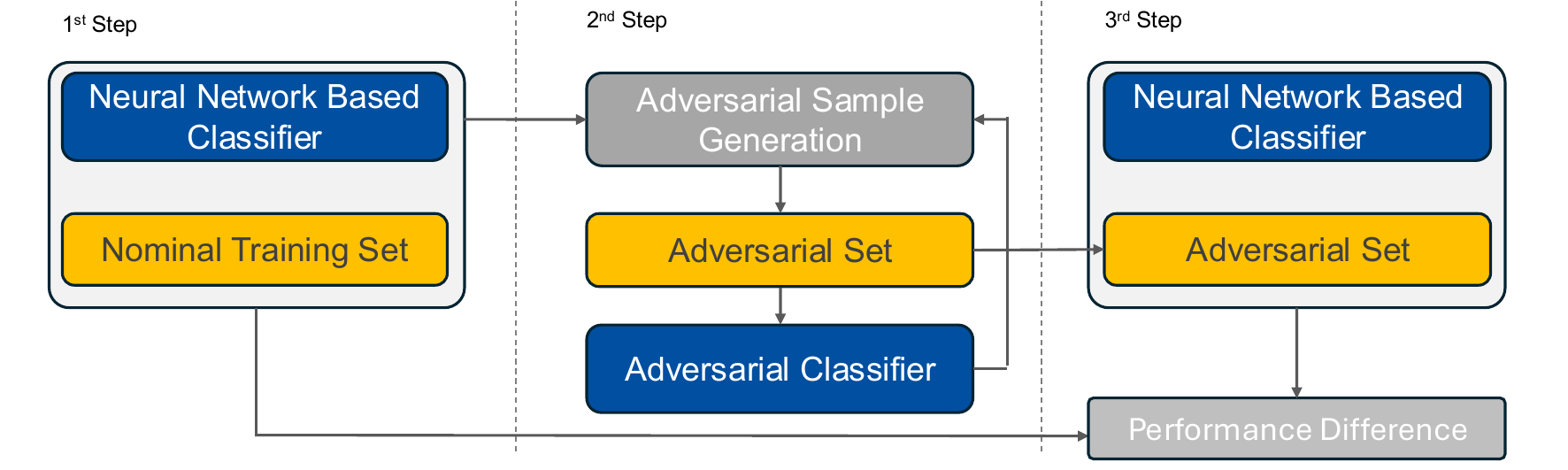}
\caption{Schematic workflow of the adversarial robustness study. Step 1: A neural network classifier is trained on the nominal training set. Step 2: Using the trained model, uncertainty-constrained adversarial samples are generated. An auxiliary adversarial classifier is trained to verify that nominal and adversarial samples are not trivially distinguishable. Step 3: The original classifier is evaluated on the adversarial dataset, and the resulting performance difference with respect to the nominal evaluation is quantified. This difference is interpreted as an estimate of the model-dependent sensitivity under experimentally allowed perturbations.\label{fig:workflow}}

\end{figure}

A practical procedure to estimate the potential size of this additional model dependence is to interpret the adversarial construction as a constrained envelope variation within a predefined uncertainty model. The suggested workflow is illustrated in Figure \ref{fig:workflow} and described in the following. After training the nominal classifier, adversarial variants of the dataset are generated using fixed, physically motivated uncertainty bounds on the input features, as described in Section~\ref{sec:framework_attack}. These bounds are chosen a priori based on detector resolutions and reconstruction uncertainties and are not tuned to maximize performance degradation.

The generated adversarial samples are required to satisfy several indistinguishability criteria. In particular, agreement of one-dimensional input feature distributions and the stability of their Pearson correlation coefficients are verified within statistical uncertainties. As an additional cross-check, an auxiliary classifier with identical architecture could be trained to distinguish nominal from adversarial samples, and its AUC is required to be statistically compatible with 0.5. These conditions ensure that the perturbations do not introduce trivially detectable distortions.

Although one-dimensional feature distributions and linear correlations remain statistically unchanged, the adversarial procedure modifies higher-order and nonlinear correlations in the joint feature space. These high-dimensional deformations are invisible in standard validation projections but can induce measurable shifts in the classifier decision boundary. The effect therefore originates from subtle geometric changes in the full feature distribution rather than from detectable distortions in individual observables.

The nominal classifier is then evaluated on the adversarial evaluation sample, and the resulting change in performance (e.g. signal efficiency at a fixed working point or ROC AUC) is quantified. This shift can be interpreted as a conservative estimate of the model dependence induced by high-dimensional variations that are consistent with experimentally allowed uncertainties but not constrained by standard validation procedures. Importantly, this estimate does not represent a guaranteed bias in nature, but rather an envelope within the assumed uncertainty model. In realistic applications, such effects are expected to be small and, ideally, subdominant to established experimental and theoretical systematic uncertainties; nevertheless, they provide a quantitative robustness check for DNN-based analyses.

\section{Conclusions and Outlook}
\label{sec:conclusions}

In this work, we have investigated the intrinsic model dependence of deep neural network classifiers in high-energy physics under physically allowed, uncertainty-constrained perturbations. We demonstrated across three representative benchmark tasks, high-level signal--background classification, graph-based quark--gluon tagging, and transformer-based $E_{\mathrm{T}}^{\mathrm{miss}}$ reconstruction, that percent-level performance degradations can occur even when one-dimensional feature distributions and linear correlations remain statistically unchanged. In contrast, simple cut-based approaches exhibit negligible sensitivity under the same conditions. These results indicate that high-capacity classifiers can be sensitive to subtle, high-dimensional deformations of the input distribution that are not captured by standard validation procedures.

Interestingly, the observed sensitivity does not scale trivially with model size: the largest relative degradation is found for the smallest architecture, suggesting that vulnerability is governed more by the structure and concentration of the learned representation than by parameter count alone. Training on combined nominal and adversarial samples substantially mitigates the observed effects, indicating that this model dependence can be reduced through explicit robustness-oriented training strategies.

The presented procedure can be interpreted as a controlled envelope test within a predefined uncertainty model. While it does not imply the presence of an actual bias in nature, it provides a quantitative robustness check for DNN-based analyses and can serve as an additional systematic sensitivity estimate complementary to conventional experimental and theoretical uncertainties.

Future work could extend this framework in several directions. A natural next step is the integration of such robustness tests directly into full analysis pipelines, including yield-level or parameter-fit impacts. Moreover, exploring alternative uncertainty models, higher-order correlation constraints, and connections to nuisance-parameter morphing may further clarify the physical interpretation of these effects. Ultimately, developing standardized robustness benchmarks for machine-learning–based analyses may help ensure that increasing model complexity is accompanied by equally rigorous uncertainty quantification.

\section*{Acknowledgements}
This work has been conducted within the context of the \emph{AISafety} and the \emph{AALearning} Project that has been funded within the funding line \emph{Software und Algorithmen zur Erforschung von Universum und Materie (ErUM) mit Schwerpunkt auf Künstlicher Intelligenz und Maschinellem Lernen} by the \emph{Bundesministerium f\"ur Bildung und Forschung (BMBF) of Germany}.

\bibliographystyle{atlasBibStyleWithTitle}
\bibliography{./Bibliography}

\end{document}